\documentclass[11pt, a4paper, logo, twocolumn, copyright, nonumbering]{googledeepmind}

\usepackage[utf8]{inputenc}
\usepackage{amssymb}
\usepackage{framed,enumitem} 
\usepackage{amsmath}
\usepackage{amsxtra}
\usepackage{mathtools}
\usepackage{graphicx}
\usepackage{hyperref}
\usepackage{bookmark}
\usepackage{booktabs}
\usepackage{nicematrix}
\usepackage{listings}
\usepackage{xcolor}
\usepackage{float}
\usepackage{makecell}
\usepackage[font=small]{caption}
\usepackage{subcaption}
\usepackage{bm}
\usepackage{siunitx}
\usepackage{braket}
\usepackage{multirow}
\usepackage{xfrac}
\usepackage{pifont}
\usepackage{cuted} 
\usepackage[super,comma,sort&compress]{natbib} 
\usepackage[normalem]{ulem} 
\definecolor{codegreen}{rgb}{0,0.6,0}
\definecolor{codegray}{rgb}{0.5,0.5,0.5}
\definecolor{codepurple}{rgb}{0.58,0,0.82}
\definecolor{backcolour}{rgb}{0.95,0.95,0.92}
\lstdefinestyle{mystyle}{
    backgroundcolor=\color{backcolour},   
    commentstyle=\color{codegreen},
    keywordstyle=\color{magenta},
    numberstyle=\tiny\color{codegray},
    stringstyle=\color{codepurple},
    basicstyle=\footnotesize,
    breakatwhitespace=false,         
    breaklines=true,                 
    captionpos=b,                    
    keepspaces=true,                 
    numbers=left,                    
    numbersep=5pt,                  
    showspaces=false,                
    showstringspaces=false,
    showtabs=false,                  
    tabsize=2
}

\definecolor{black}{rgb}{0,0,0} 
\definecolor{darkgreen}{rgb}{0,0.6,0}
\newcommand{\rebuttal}[1]{{\textcolor{black}{#1}}}

\newcommand{\wrt}{w.\,r.\,t.~}

\newcommand\T{{\mathpalette\raiseT\intercal}}
\newcommand\raiseT[2]{\raisebox{0.25ex}{$#1#2$}}
\newcommand{\compconj}[1]{
  \overline{#1}
}

\newcommand{\xmark}{\ding{55}}

\DeclareMathOperator*{\Motimes}{\text{\raisebox{0.25ex}{\scalebox{0.8}{$\bigotimes$}}}} 

\newcommand{\numgridpoints}{G}

\newcommand{\an}{\si{\angstrom}}

\providecommand\phantomsection{}

\hypersetup{bookmarksdepth=4}

\title{Euclidean Fast Attention -- Machine Learning Global Atomic Representations at Linear Cost}

\correspondingauthor{oliverunke@google.com}

\reportnumber{} 

\renewcommand{\today}{}

\author[1, 2, 3]{J.~Thorben Frank}
\author[2, 3]{Stefan Chmiela}
\author[1, 2, 3, 4, 5]{Klaus-Robert Müller}
\author[1]{Oliver T.~Unke}

\affil[1]{Google DeepMind, Berlin, Germany}
\affil[2]{Machine Learning Group, Technische Universit\"at Berlin, 10587 Berlin, Germany}
\affil[3]{Berlin Institute for the Foundations of Learning and Data -- BIFOLD, Germany}
\affil[4]{Max Planck Institute for Informatics, Stuhlsatzenhausweg, 66123 Saarbr{\"u}cken, Germany}
\affil[5]{Department of Artificial Intelligence, Korea University, Anam-dong, Seongbuk-gu, Seoul 02841, Korea}

\begin{abstract}
Long-range correlations are essential across numerous machine learning tasks, especially for data embedded in Euclidean space, where the relative positions and orientations of distant components are often critical for accurate predictions. Self-attention offers a compelling mechanism for capturing these global effects, but its \emph{quadratic} complexity presents a significant practical limitation. This problem is particularly pronounced in computational chemistry, where the stringent efficiency requirements of machine learning force fields (MLFFs) often preclude accurately modelling long-range interactions. To address this, we introduce Euclidean fast attention (EFA), a \emph{linear}-scaling attention-like mechanism designed for Euclidean data, which can be easily incorporated into existing model architectures. A core component of EFA are novel Euclidean rotary positional encodings (ERoPE), which enable efficient encoding of spatial information while respecting essential physical symmetries. We empirically demonstrate that EFA effectively captures diverse long-range effects, enabling EFA-equipped MLFFs to describe challenging chemical interactions for which conventional MLFFs yield incorrect results.
\end{abstract}
\begin{document}

\maketitle

\renewcommand{\figureautorefname}{Fig.}
\renewcommand{\equationautorefname}{Eq.}
\renewcommand{\tableautorefname}{Tab.}
\renewcommand{\thesubfigure}{\Alph{subfigure}}

\phantomsection
\section{Introduction}
\label{sec:introduction}

Many applications of machine learning (ML) are characterised by a complex interplay of short-range (local) and long-range (global) correlations. modelling global correlations is particularly challenging for data embedded in Euclidean space, where the relationships between distant components can be governed by both their relative positions \emph{and} orientations. For applications of ML in physics and computational chemistry,\cite{rupp2012fast,von2020exploring,unke2021machine, keith2021combining} e.g., for performing molecular dynamics (MD) simulations, matters are further complicated by the fact that atomic interactions obey certain symmetry relations (e.g., translational/rotational invariance/equivariance), and violating these constraints can severely degrade performance.\cite{fu2022forces}

MD simulations allow to study the movement of individual atoms over time and provide insights into the structure, function, and thermodynamics of molecular systems and materials.\cite{karplus2002molecular} The accuracy of such simulations crucially depends on the quality of the description of the forces between atoms, which drive their motion. An increasingly popular method for modelling these interactions are machine learning force fields (MLFFs),\cite{behler2007generalized, bartok2010gaussian, schutt2018schnet,schutt2020machine,unke2021machine,keith2021combining} which can reach the accuracy of \textit{ab initio} electronic structure calculations at a fraction of the computational cost.
MLFFs are now routinely used to study a variety of complex systems and phenomena, with applications spanning from protein dynamics\cite{unke2024biomolecular, kabylda2024molecular} to the discovery of new materials.\cite{merchant2023scaling} However, despite remarkable progress in the development of MLFFs in recent years, accurately modelling long-range effects with MLFFs remains a persistent challenge. This shortcoming often stems from stringent computational efficiency requirements: Many systems of practical interest (e.g., bio-molecules) consist of hundreds of thousands of atoms, so linear scaling \wrt the number of atoms is a prerequisite. The lack of an accurate treatment of long-range interactions is problematic, because although they are usually comparatively weak (in magnitude), they can play a crucial role for the stability, long time-scale dynamics, structure, and response properties of a variety of chemical and biological systems.\cite{woods2016materials,hermann2017first,stohr2019theory}

In principle, the self-attention mechanism underlying the transformer architecture\cite{vaswani2017attention} offers a compelling mechanism for capturing such global effects. It has revolutionised natural language processing and, due to its flexibility and generality, also found applications in many other fields, including computer vision,\cite{dosovitskiy2020image} graph learning,\cite{velivckovic2017graph} and the natural sciences.\cite{von2022self} Unfortunately, standard self-attention has quadratic time and memory cost in the number of inputs, hindering its wide-scale adoption in MLFFs. While the memory cost of self-attention can be reduced to scale linearly,\cite{dao2022flashattention, dao2023flashattention} the time complexity remains quadratic. 
\rebuttal{\label{rebuttal:efficient_attention_formulations} 
To overcome this limitation, many alternative formulations\cite{tay2022efficient} and algorithmic modifications\cite{yuan2025native} of attention have been proposed, including linear-scaling variants.\cite{katharopoulos2020transformers, choromanski2020rethinking} However, there is a fundamental issue that prevents applying these techniques to data embedded in Euclidean space: When modelling interatomic forces, the relative spatial arrangement of atoms is crucially important. While geometric information can be easily encoded in standard quadratic-scaling attention using pair-wise terms, it is unclear how to do so in linear-scaling formulations. Encoding all relevant geometric information at linear complexity is non-trivial and requires specialised methods (see SI \nameref{sec:attention-and-euclidean-data} for additional context).}

In this work, we address these challenges by proposing the linear-scaling Euclidean fast attention (EFA) mechanism. It enables learning global representations that encode the spatial structure of a chemical system while respecting all relevant physical symmetries. This is achieved via our novel Euclidean rotary positional encodings (ERoPE), which allow a description of the relative positions and orientations of atoms with linear complexity. EFA can be incorporated into existing local MLFFs with minimal architectural modifications, enabling them to accurately model global correlations. We empirically demonstrate that EFA-augmented models are able to describe various long-range interactions and non-local effects, while MLFFs without EFA yield incorrect results as they are unable to capture long-range structure.

\rebuttal{
\label{rebuttal:shortened_introduction} 
\subsection{Limitations of local models}\label{subsec:limitations-of-local-models}
Before presenting our results, we briefly review the limitations of local MLFFs with respect to modelling long-range interactions. Throughout this work, we focus on message passing neural networks (MPNNs),\cite{gilmer2017neural} a popular model class for constructing MLFFs,\cite{zhang2018end,unke2021spookynet,batzner20223,frank2024euclidean,poltavsky2024crash1,poltavsky2024crash2} to demonstrate the difference between conventional (local) and EFA-augmented models. However, our findings are also relevant for most other types of local MLFFs, which typically have similar locality properties (see below). MPNNs represent chemical structures as graphs embedded in Euclidean space, where individual atoms correspond to nodes with associated features $\mathcal{X} = \{\boldsymbol{x}_1,\dots,\boldsymbol{x}_N\,|\,\boldsymbol{x}_m\in\mathbb{R}^H\}$ and positions $\mathcal{R} = {\{\vec{r}_1, \dots, \vec{r}_N\,|\,\vec{r}_m \in \mathbb{R}^3\}}$. Starting from the initial features $\mathcal{X}^{[0]}$, they are updated via $T$ message passing (MP) layers, where all nodes connected by an edge ``pass messages'' to each other. The final representations $\mathcal{X}^{[T]}$ are then used to predict atom-wise energy contributions ${\{E_1, \dots, E_N\,|\,E_m \in \mathbb{R}\}}$.}

\rebuttal{Evaluating an MPNN scales as $\mathcal{O}(N^2)$ for fully-connected graphs. To achieve $\mathcal{O}(N)$ scaling, most MPNNs introduce a \emph{local cutoff} $r_\text{cut}$, and two nodes $m$ and $n$ are only considered connected (``neighbours'') to each other if $\lVert\vec{r}_m - \vec{r}_n\rVert < r_\text{cut}$. \label{rebuttal:more_visible_allegro_citation}Other MLFF model architectures typically employ similar cutoff strategies. For example, they may rely on fixed-size descriptors of local atomic environments, which are used as inputs to a feed-forward neural network\cite{smith2017ani} or within kernel methods.\cite{christensen2020fchl} A similar construction is used in Allegro,\cite{musaelian2023learning} where local descriptors are formed via tensor products of learned equivariant representations that depend on neighbouring atoms. While a cutoff is desirable from an efficiency perspective, it also means that interactions beyond a certain distance, which we refer to as \emph{effective cutoff}, cannot be modelled by construction.}

\begin{figure*}[p!]
    \centering
    \includegraphics[width=\textwidth   ]{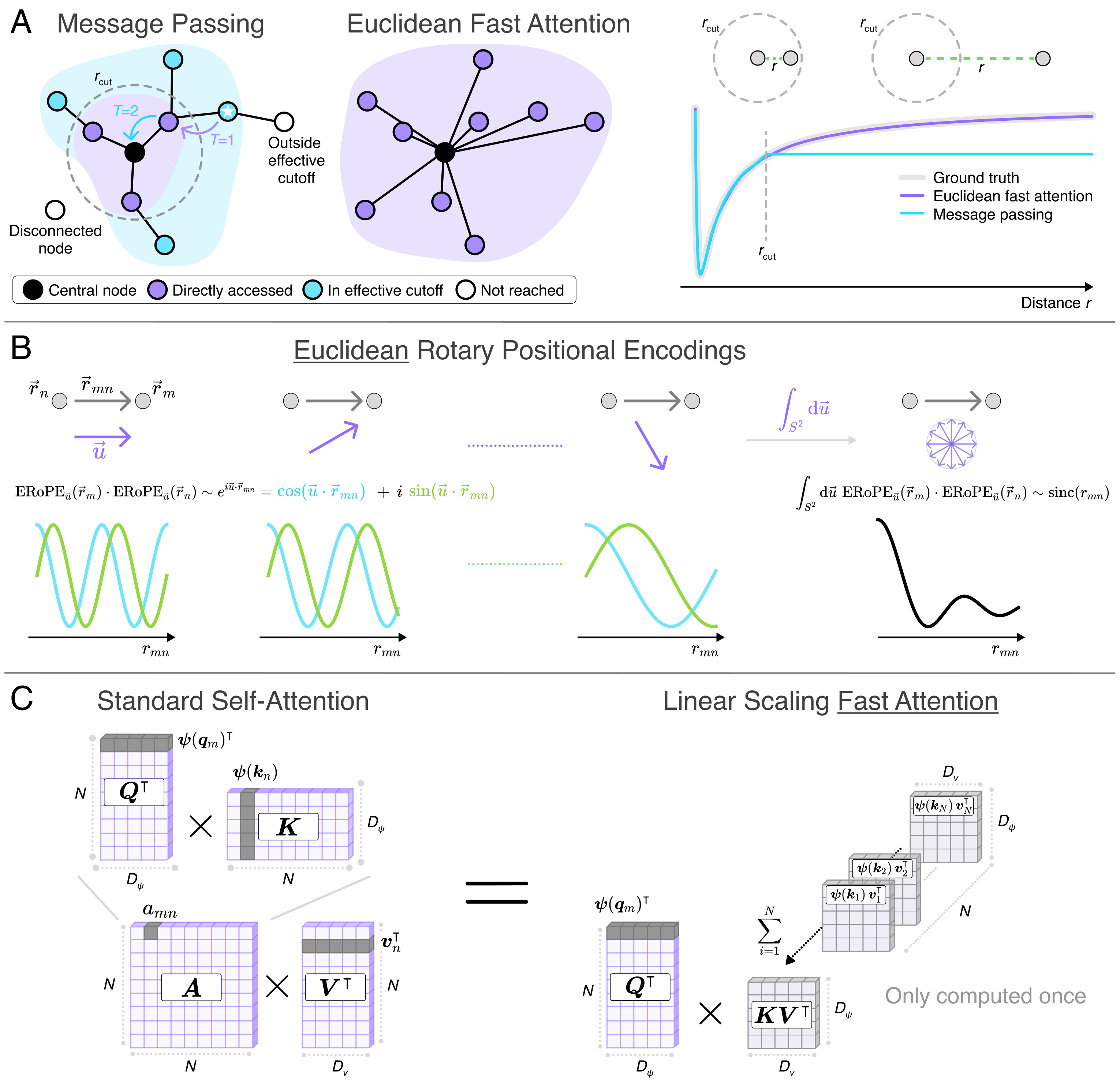}
    \begin{subfigure}{\linewidth}
        \phantomcaption{}
        \label{fig:overview:mp-vs-efa-information-flow-and-pairwise-potential}
    \end{subfigure}
    \begin{subfigure}{\linewidth}
        \phantomcaption{}
        \label{fig:overview:erope}
    \end{subfigure}
    \begin{subfigure}{\linewidth}
        \phantomcaption{}
        \label{fig:overview:quadratic-vs-linear-attention}
    \end{subfigure}
    \caption{\textbf{Overview of the central concepts of this work.} \textbf{(A)} Information about the overall structure of a graph/molecule accessible by a central node/atom. With message passing (MP), only nodes within the local cutoff $r_\text{cut}$ (purple) can directly pass information to the central node (black). With $T=2$ update blocks (\autoref{eq:mp-update-block}), information about more distant nodes (cyan) also becomes available (the arrows illustrate how information ``hops'' from the highlighted node ($\star$) to the central node). Unreachable nodes (white) are either outside the effective cutoff $T\cdot r_\text{cut}$, or disconnected (there is no ``hopping path''). In contrast,  with Euclidean fast attention (EFA), all nodes can be accessed directly -- irrespective of distance. This enables EFA to faithfully capture long-range effects, illustrated here for a simple pairwise potential $V(r)$, where MP fails to model $V(r)$ when $r$ is larger than the cutoff. \textbf{(B)} EFA encodes geometric information using Euclidean rotary positional encodings (ERoPE), see \autoref{eq:euclidean-rotary-positional-encoding}. ERoPE effectively projects the displacement vectors $\vec{r}_{mn}$ between nodes $m$ and $n$ onto a unit vector $\vec{u}$, and expands the result in a complex exponential $e^{i \vec{u}\cdot\vec{r}_{mn}}$. The frequency of the resulting ``wave'' depends on the angle between $\vec{r}_{mn}$ and the chosen $\vec{u}$. By averaging over all possible choices of $\vec{u}$ (integration over the unit sphere $S^2$), a rotationally invariant encoding (independent of $\vec{u}$) of the distance $r_{mn}$ can be obtained. \textbf{(C)} Schematic representation of (quadratically-scaling) standard self-attention (\autoref{eq:linear-scaling-attention-still-quadratic}) vs.\ a linear-scaling formulation\cite{katharopoulos2020transformers} (\autoref{eq:linear-scaling-attention}). Here, the operation is shown in matrix form, which computes attention for all inputs at once.}
    \label{fig:overview}
\end{figure*}

\rebuttal{For MLFFs based on strictly local descriptors, the effective cutoff is equal to $r_\text{cut}$, but MPNNs are special in that the effective cutoff can exceed $r_\text{cut}$.} This is because nodes can also gather information from other nodes they are not directly connected to, as long as this information could first propagate to one of their direct neighbours in previous MP layers. The maximally possible effective cutoff is $T\cdot r_\text{cut}$, but if there is no ``hopping path'' between two nodes to act as ``relay'' over multiple updates, they cannot exchange information, even if their distance is below $T\cdot r_\text{cut}$ (see ``disconnected node'' in \autoref{fig:overview:mp-vs-efa-information-flow-and-pairwise-potential}). Further, this indirect information transfer between nodes only captures a ``mean-field effect'', often insufficient to accurately capture long-range interactions (see \nameref{sec:results}).

\phantomsection
\section{Results}
\label{sec:results}
In the following, we address the challenge of modelling global interactions with linear complexity by proposing Euclidean Fast Attention (EFA). We then contrast EFA with MP on idealised model systems to highlight important differences in their properties and their ability to model global correlations under controlled conditions. Finally, we apply EFA-augmented and standard MPNN architectures to several challenging realistic chemical systems and show that including EFA improves the description of long-range interactions significantly.

\subsection{Euclidean fast attention}
\label{sec:euclidean-fast-attention}
Inspired by rotary positional encodings (RoPE),\cite{su2024roformer} we propose a mechanism to encode positions $\vec{r} \in \mathbb{R}^3$ in Euclidean space into feature vectors $\boldsymbol{x}$, which we call Euclidean RoPE (ERoPE):
\begin{equation}
\mathrm{ERoPE}_{\vec{u}}(\boldsymbol{x}, \vec{r}) \coloneqq \boldsymbol{x}\cdot e^{i \omega \vec{u}\cdot \vec{r}}\,.
\label{eq:euclidean-rotary-positional-encoding}
\end{equation}
Here, $i$ is the imaginary unit, $\omega \in \mathbb{R}$ is a frequency coefficient, $\vec{u}\cdot \vec{r}$ is the dot product of $\vec{u}$ and $\vec{r}$, and $\vec{u} \in S^2$ is a three-dimensional unit vector. Within the context of attention mechanisms, after encoding positions $\vec{r}_m$ and $\vec{r}_n$ with \autoref{eq:euclidean-rotary-positional-encoding} into queries $\boldsymbol{q}$ and keys $\boldsymbol{k}$, their scalar product (a core component of attention, see SI \nameref{sec:attention-and-euclidean-data}) is
\begin{equation*}
\begin{aligned}
\langle\boldsymbol{q}\cdot e^{i \omega \vec{u}\cdot \vec{r}_m}, \boldsymbol{k}\cdot e^{i \omega \vec{u}\cdot \vec{r}_n}\rangle 
&= \left(\boldsymbol{q}\cdot e^{i \omega \vec{u}\cdot \vec{r}_m}\right)^\T \left(\compconj{\boldsymbol{k}\cdot e^{i \omega \vec{u}\cdot \vec{r}_n}}\right)\\
&= \boldsymbol{q}^\T\compconj{\boldsymbol{k}} \cdot e^{i \omega \vec{u}\cdot \vec{r}_m} \compconj{e^{i \omega \vec{u}\cdot \vec{r}_n}}\\
&=
\langle\boldsymbol{q},\boldsymbol{k}\rangle \cdot e^{i \omega \vec{u}\cdot(\vec{r}_m-\vec{r}_n)}\,.
\end{aligned}
\end{equation*}
Since $\vec{u}\cdot(\vec{r}_m-\vec{r}_n)$ is the projection of the vector $\vec{r}_{mn} = \vec{r}_m-\vec{r}_n$ onto $\vec{u}$, the scalar product now contains geometric information about the relative displacement vector $\vec{r}_{mn}$.

Unfortunately, while displacement vectors are invariant with respect to translations, the value of the projection $\vec{u}\cdot\vec{r}_{mn}$ depends on the choice of $\vec{u}$ and is therefore \emph{not} invariant with respect to rotations. This issue can be resolved by integrating over $S^2$ (see \autoref{fig:overview:erope}):
\begin{equation}
\frac{1}{4\pi}\int_{S^2} e^{i \omega \vec{u}\cdot\vec{r}_{mn}} \ \mathrm{d}\vec{u} = \frac{\sin(\omega r_{mn})}{\omega r_{mn}}=\mathrm{sinc}(\omega r_{mn})\,.
\label{eq:surface-integral-solution}
\end{equation}
The resulting expression only depends on the relative distance $r_{mn} = \lVert\vec{r}_{mn}\rVert$ and is therefore invariant with respect to rotations (see SI \nameref{si:sec:surface-integral}). It is also possible to encode different components with varying frequencies (corresponding to a mixture of multiple $\mathrm{sinc}$ functions), which improves the expressive power of the operation (see \nameref{sec:methods}).

\rebuttal{\label{rebuttal:efa_extension_to_pbcs}
We note that the integration in \autoref{eq:surface-integral-solution} is only necessary for applications where invariance w.r.t.\ $\mathrm{SO}(3)$ is a strict requirement. When describing materials under periodic boundary conditions (PBCs), symmetry w.r.t.\ $\mathrm{SO}(3)$ is broken within the unit cell. In this case, an alternative to \autoref{eq:surface-integral-solution} is picking special directions for $\vec{u}$ (e.g., the lattice vectors). This variant is discussed in more detail in SI \nameref{si:sec:other-symmetrisation-operations}. In this context, the general form of Eq.~\ref{eq:euclidean-rotary-positional-encoding} resembles a plane-wave expansion and therefore shares some similarities with other approaches that evaluate interactions in the frequency domain, such as methods based on Ewald summation (see SI \nameref{si:sec:relation-to-ewald-mp}).\cite{ewald1921berechnung,kosmala2023ewald}}

We now combine ERoPE, integration over $S^2$, and ideas from linear-scaling attention to arrive at Euclidean fast attention (EFA). It is given by
\begin{equation}
\begin{aligned}
    \mathrm{EFA}&(\mathcal{X}, \mathcal{R})_m =\\ &\frac{1}{4\pi}\int_{S^2} \boldsymbol{\phi}_{\vec{u}}(\boldsymbol{q}_m, \vec{r}_m)^\T\sum_{n=1}^N  \compconj{\boldsymbol{\phi}_{\vec{u}}(\boldsymbol{k}_n, \vec{r}_n)} \, \boldsymbol{v}_n^\T \ \mathrm{d}\vec{u}\,,
\end{aligned}
\label{eq:euclidean-fast-attention}
\end{equation}
where we define the short-hand $\boldsymbol{\phi}_{\vec{u}}(\boldsymbol{x}, \vec{r}) \coloneqq \mathrm{ERoPE}_{\vec{u}}(\boldsymbol{\psi}(\boldsymbol{x}), \vec{r})$ for conciseness and $\boldsymbol{\psi}$ can be any feature map (see \autoref{eq:similarity_kernel_rewrite}). We base EFA on an \emph{attention-like} mechanism given by
\begin{equation}
\widetilde{\mathrm{ATT}}_{\text{Lin}}(\mathcal{X})_m = \boldsymbol{\psi}(\boldsymbol{q}_m)^\T\sum_{n=1}^N  \boldsymbol{\psi}(\boldsymbol{k}_n) \, \boldsymbol{v}_n^\T\,,
\label{eq:linear-scaling-attention-no-denominator}
\end{equation}
which omits the denominator compared to standard attention formulations (see \autoref{eq:linear-scaling-attention}). Omitting normalisation is natural when modelling long-range interactions in chemical systems since their effects are typically additive (size extensive).

Finally, there is an additional generalisation we can make, which further increases the geometric expressiveness of \autoref{eq:euclidean-fast-attention}. So far, we have implicitly assumed that the features from which queries, keys, and values are computed are themselves rotationally invariant. However, recent MLFFs often use \emph{equivariant features},\cite{batzner20223,frank2022so3krates} which can be thought of as containing additional 
``directional information''.\cite{unke2024e3x} We can generalise the EFA mechanism as
\begin{equation}\label{eq:euclidean-fast-attention-equivariant}
\begin{aligned}
&\mathrm{EFA}(\mathcal{X}, \mathcal{R})_m =\\ 
&\hspace{10pt}\frac{1}{4\pi}\!\int_{S^2}\! \boldsymbol{\phi}_{\vec{u}}(\boldsymbol{q}_m, \vec{r}_m)^\T\sum_{n=1}^N  \compconj{\boldsymbol{\phi}_{\vec{u}}(\boldsymbol{k}_n, \vec{r}_n)} \, \boldsymbol{v}_n^\T\otimes\boldsymbol{Y}({\vec{u}})\ \mathrm{d}\vec{u}\,,
\end{aligned}
\end{equation}

so it is also applicable to equivariant features. Here, `$\otimes$' denotes a tensor product and $\boldsymbol{Y}(\vec{u})$ is a $(\ell_\text{max}+1)^2$--vector containing all \emph{spherical harmonics} up to degree $\ell_\text{max}$. For invariant input features and $\ell_\text{max}=0$, \autoref{eq:euclidean-fast-attention-equivariant} simplifies to \autoref{eq:euclidean-fast-attention}. 
The resulting atomic features are of the form 
\begin{align}
    \boldsymbol{x}_m \simeq \sum_{n=1}^N \boldsymbol{f}(r_{mn}) \circ  \boldsymbol{x}_n \otimes \boldsymbol{Y}(\hat{r}_{mn}),
    \label{eq:so3-convolutions-main}
\end{align}
where $\boldsymbol{f}$ is a vector-valued radial function and `$\circ$' denotes element-wise multiplication (see SI \nameref{si:sec:surface-integral}). \autoref{eq:so3-convolutions-main} resembles the structure of $\mathrm{SO}(3)$ convolutions -- the basic MP building block in many equivariant MPNNs \cite{liao2022equiformer,batzner20223} -- but now {\em without} needing to introduce a local cutoff to achieve linear scaling.

\begin{figure*}[t!]
    \centering
    \includegraphics[width=\linewidth]{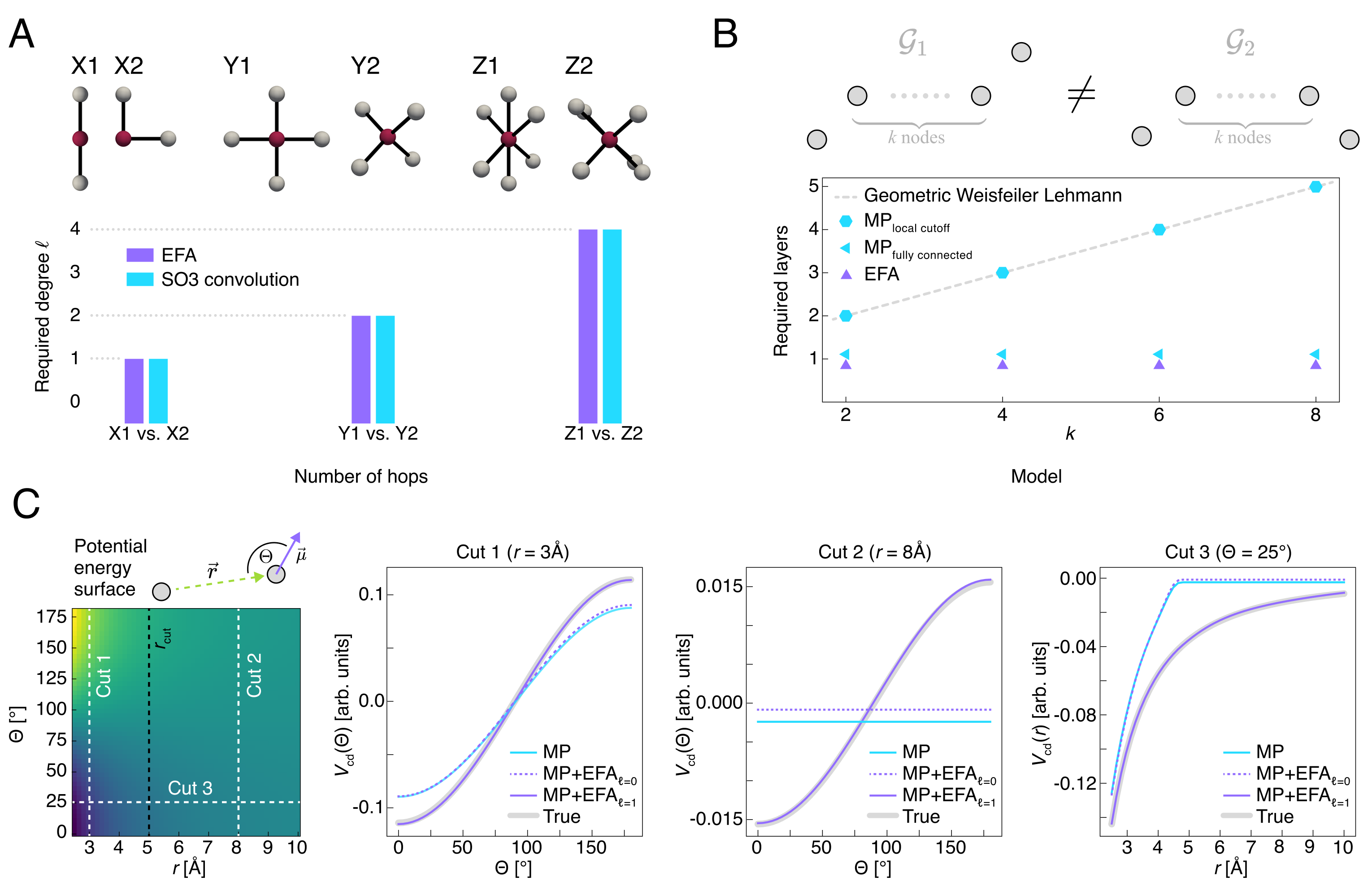}
    \begin{subfigure}{\linewidth}
        \phantomcaption{}
        \label{fig:geometric-expressiveness:environment-pairs}
    \end{subfigure}
    \begin{subfigure}{\linewidth}
        \phantomcaption{}
        \label{fig:geometric-expressiveness:weisfeiler-leman}
    \end{subfigure}
    \begin{subfigure}{\linewidth}
        \phantomcaption{}
        \label{fig:results-panel-idealised-systems:dipole-potential}
    \end{subfigure}
    \caption{\textbf{Geometric expressiveness of Euclidean fast attention (EFA).} \textbf{(A)} Smallest degree $\ell$ of spherical harmonics $\boldsymbol{Y}$ required for EFA (\autoref{eq:euclidean-fast-attention-equivariant}) and  $\mathrm{SO}(3)$ convolutions (\autoref{eq:so3-convolutions-main}) to distinguish pairs (X1/X2, Y1/Y2, and Z1/Z2) of distinct environments\cite{pozdnyakov2020incompleteness} of the central red node (in a single MP/update). None of the pairs can be distinguished by using invariant ($\ell = 0$) information only (unless multiple updates are used), because all neighbours are equidistant from the central node. \textbf{(B)} Number of MP layers required to distinguish the graphs $\mathcal{G}_1$ and $\mathcal{G}_2$ for different chain lengths $k$.\cite{joshi2023expressive} The graphs are non-isomorphic according to a geometric Weisfeiler-Leman\cite{leman1968reduction} (GWL) test, due to different orientations of the terminal nodes \wrt each other. With MP (using a local cutoff), at least $T=\lfloor k/2\rfloor + 1$ layers are required to solve the task, following the theoretically expected behaviour from the GWL test (grey dashed line). In contrast, with EFA, $\mathcal{G}_1$ and $\mathcal{G}_2$ can always be distinguished with a single update (similar to MP on a fully connected graph). \textbf{(C)} Standard MP and MP+EFA$_\ell$ applied to a non-isotropic interaction ($\ell$ refers to the maximal degree of the spherical harmonics vector $\mathbf{Y}$ in \autoref{eq:euclidean-fast-attention-equivariant}). Energy predictions along different cuts of the two-dimensional potential energy surface (see leftmost panel for reference) are shown. Only MP+EFA$_{\ell=1}$ has access to information about both relevant degrees of freedom (distance and orientation) and can therefore faithfully describe the interaction beyond the cutoff $r_{\text{cut}}$.}
    \label{fig:geometric-expressiveness}
\end{figure*}

\phantomsection
\subsection{Idealised systems}
\label{sec:idealised-systems}
Before applying EFA to realistic molecular systems \rebuttal{and materials}, we evaluate and contrast it with standard MP on idealised model systems.

\phantomsection
\subsubsection{Geometric expressiveness}
\label{sec:geometric-expressiveness}
First, we investigate the geometric expressiveness of EFA and MP, i.e., their ability to resolve geometric information. When distinguishing increasingly complex pairs of local atomic neighbourhoods,\cite{pozdnyakov2020incompleteness} we empirically confirm the similarity between EFA and $\mathrm{SO}(3)$ convolutions and find that increasing the degree $\ell$ in $\boldsymbol{Y}_\ell$ has the same effect for both operations (see \autoref{fig:geometric-expressiveness:environment-pairs} and \nameref{sec:methods}). 

To analyse their ability of distinguishing molecular graphs as a whole, we test whether MP and EFA can distinguish two graphs that are non-isomorphic according to a geometric Weisfeiler-Leman\cite{leman1968reduction, joshi2023expressive} test (see \autoref{fig:geometric-expressiveness:weisfeiler-leman}). Unlike a standard MP model, which requires an increasing number of layers to distinguish the non-isomorphic graphs, EFA accomplishes this within a single update by capturing global structural information.

\subsubsection{Pairwise Potentials}
Next, we consider pairwise potentials, such that the total energy of a system of $N$ atoms is given by
\begin{align}
    E(\vec{r}_1, \dots, \vec{r}_N) = \sum_{m}^N\sum_{n>m}^N V(\lVert\vec{r}_m - \vec{r}_n\rVert),
    \label{eq:pairwise-potential}
\end{align}
where, $V$ is a scalar potential function.

\begin{figure*}[t!]
    \centering
    \includegraphics[width=\linewidth]{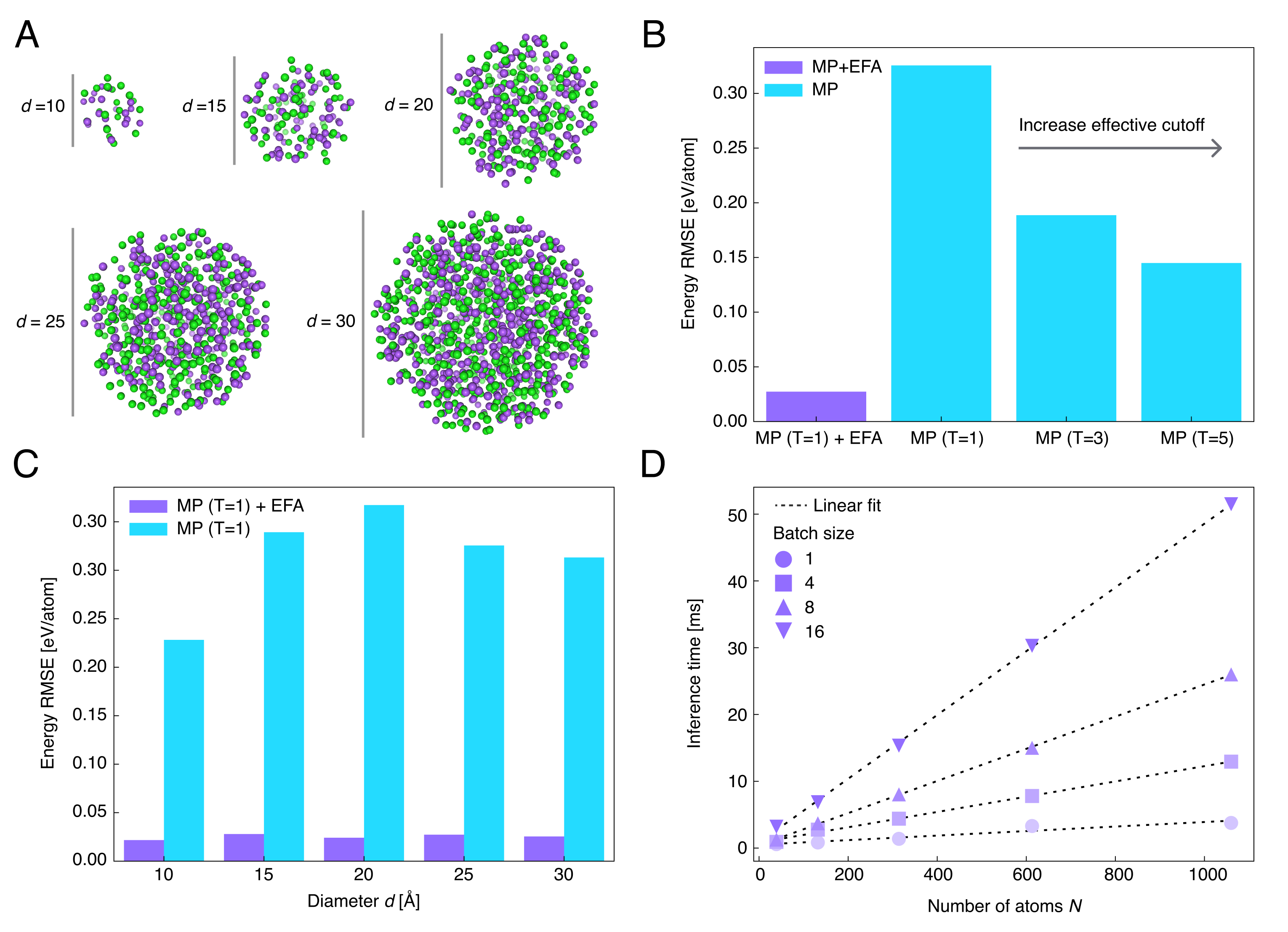}
    \begin{subfigure}{\linewidth}
        \phantomcaption{}
        \label{fig:results-panel-NaCl-toy:structures}
    \end{subfigure}
    \begin{subfigure}{\linewidth}
        \phantomcaption{}
        \label{fig:results-panel-NaCl-toy:mp-increase-T}
    \end{subfigure}
    \begin{subfigure}{\linewidth}
        \phantomcaption{}
        \label{fig:results-panel-NaCl-toy:energy-rmse-per-atom-vs-size-mp-efa}
    \end{subfigure}
    \begin{subfigure}{\linewidth}
        \phantomcaption{}
        \label{fig:results-panel-NaCl-toy:number-of-atoms-vs-time}
    \end{subfigure}
    \caption{\rebuttal{\textbf{Shortcomings of Message Passing (MP) and Scaling Analysis.} \textbf{(A)} Visual illustration of NaCl-like test systems with increasing diameter $d$ and number of atoms. \textbf{(B)} Root mean square error (RMSE) of energy predictions for a single layer MP model augmented with Euclidean fast attention (EFA) and a standard MP model with an increasing number of layers for the $d=25\,\an$ system (all variants use a cutoff of $5\,\an$). \textbf{(C)} Energy RMSE for different NaCl systems for a single layer MP model and an EFA-augmented variant. \textbf{(D)} Inference time for MP(T=1)+EFA as function of the number of atoms (see panel A) for different batch sizes (dashed lines show linear fits). 
    }}
    \label{fig:results-panel-NaCl_toy}
\end{figure*}

\paragraph{Two-Particle Systems}
We start with the simplest case of two particles ($N=2$) and choose $V$ such that the long-range decay behaviour is reminiscent of (attractive) charge-charge interactions (see \nameref{sec:methods}). As expected, a standard MPNN predicts a qualitatively wrong, constant energy profile as soon as the particle separation exceeds the cutoff distance. When the model is augmented with EFA, the pairwise potential can be accurately described over the full interaction length (see \autoref{fig:overview:mp-vs-efa-information-flow-and-pairwise-potential}). 

To reveal the shortcomings of purely distance-based descriptions and show the utility of the equivariant version of the EFA mechanism (\autoref{eq:euclidean-fast-attention-equivariant}), the potential $V$ is made to mimic a  non-isotropic charge-dipole interaction, i.e., it now not only depends on distance, but also orientation (see \nameref{sec:methods}). Only EFA models with $\ell \geq 1$ in the spherical harmonics vector $\boldsymbol{Y}_{\ell}$ capture the true nature of the underlying potential over all length scales (Fig.~\ref{fig:results-panel-idealised-systems:dipole-potential}).

\paragraph{\rebuttal{N-Particle Systems}}
\rebuttal{\label{rebuttal:n-particle-system} Next we consider increasingly large NaCl-like systems (\autoref{fig:results-panel-NaCl-toy:structures}), where negatively charged Cl and positively charged Na ions interact via a screened Coulomb potential (see \nameref{sec:methods}).}

\rebuttal{For each system, we train a single layer MP model with and without EFA block. We find, that employing EFA strongly improves the accuracy (\autoref{fig:results-panel-NaCl-toy:energy-rmse-per-atom-vs-size-mp-efa}) and scales linear in the number of atoms (\autoref{fig:results-panel-NaCl-toy:number-of-atoms-vs-time}).} 

\rebuttal{We further compare MP+EFA to MPs with an increasing number of layers, such that the effective cutoff covers the whole system (\autoref{fig:results-panel-NaCl-toy:mp-increase-T}). Our results suggest that ``mean-field interactions'' between node neighbourhoods in MPNNs do not capture all relevant details of interatomic potentials, and augmenting models with EFA is still beneficial in this setting.}

\phantomsection
\subsection{\rebuttal{Molecular Systems and Materials}}
\rebuttal{In the following, we investigate the performance of EFA on realistic data. We study several atomistic systems exhibiting various long-ranged and non-local effects, which are representative for common interaction patterns found throughout chemistry. We demonstrate that EFA systematically improves the performance of local MPNNs and that an accurate description of long-range interactions can be crucial for predicting the dynamic properties of molecular systems.
\paragraph{Non-local charge transfer}
\label{rebuttal:long-range-benchmark} Ko.\,et~al.\, proposed a benchmark for analysing the capability of MLFFs to model non-local charge transfer effects.\cite{ko2021fourth} The test systems include molecules and materials, requiring applicability to systems with and without periodic boundary conditions (PBCs). For each system we train a regular MPNN and an EFA-augmented variant (see \nameref{sec:methods}) and compare energy and force errors to other state-of-the-art methods (see \autoref{tab:long-range-bechmark}). For the standard MPNN we see increased errors, whereas the model with EFA outperforms all other methods in 7 out of 8 metrics.}
\begin{table*}[t!]
\centering
\begin{tabular}{llcccccc}
\toprule
\textbf{System} & \textbf{Metric} & \textbf{PBC} & \textbf{2G-BPNN} & \textbf{4G-BPNN} & \textbf{SpookyNet} & \textbf{MP} & \textbf{MP+EFA} \\ 
\midrule
\multirow{2}{*}{Au$_2$-MgO} & \textit{Energy} & \multirow{2}{*}{\checkmark} & 2.287 & 0.219 & 0.107 & 2.225 & \textbf{0.088} \\
 & \textit{Forces} & & 153.100 & 66.000 & 5.337 & 55.310 & \textbf{4.230} \\
\midrule
\multirow{2}{*}{Na$_{8/9}$Cl$_8^+$} & \textit{Energy} & \multirow{2}{*}{\xmark} & 1.692 & 0.481 & 0.135 & 1.485 & \textbf{0.109} \\
 & \textit{Forces} & & 57.390 & 32.780 & \textbf{1.052} & 34.383 & 1.205 \\
\midrule
\multirow{2}{*}{C$_{10}$H$_2$\,/\,C$_{10}$H$_3^+$} & \textit{Energy} & \multirow{2}{*}{\xmark} & 1.619 & 1.194 & 0.364 & 0.704 & \textbf{0.254} \\
 & \textit{Forces} & & 129.500 & 78.000 & 5.802 & 29.608 & \textbf{5.351} \\
\bottomrule
\end{tabular}
\caption{\rebuttal{\textbf{Non-local charge transfer benchmark.} Root mean square error (RMSE) for energy (meV/atom) and forces (meV/$\si{\angstrom}$) on the benchmark introduced in Ref.~\citenum{ko2021fourth}. Euclidean fast attention (EFA) can be applied to systems with and without periodic boundary conditions (PBCs), surpassing local models (our MP, 2G-BPNN\cite{behler2007generalized}) by a large margin. It also outperforms other non-local approaches (4G-HDNNP~\cite{ko2021fourth}, SpookyNet~\cite{unke2021spookynet}). Results are averaged over two runs.}}
\label{tab:long-range-bechmark}
\end{table*}
\begin{figure*}[t!]
    \centering
    \includegraphics[width=\linewidth]{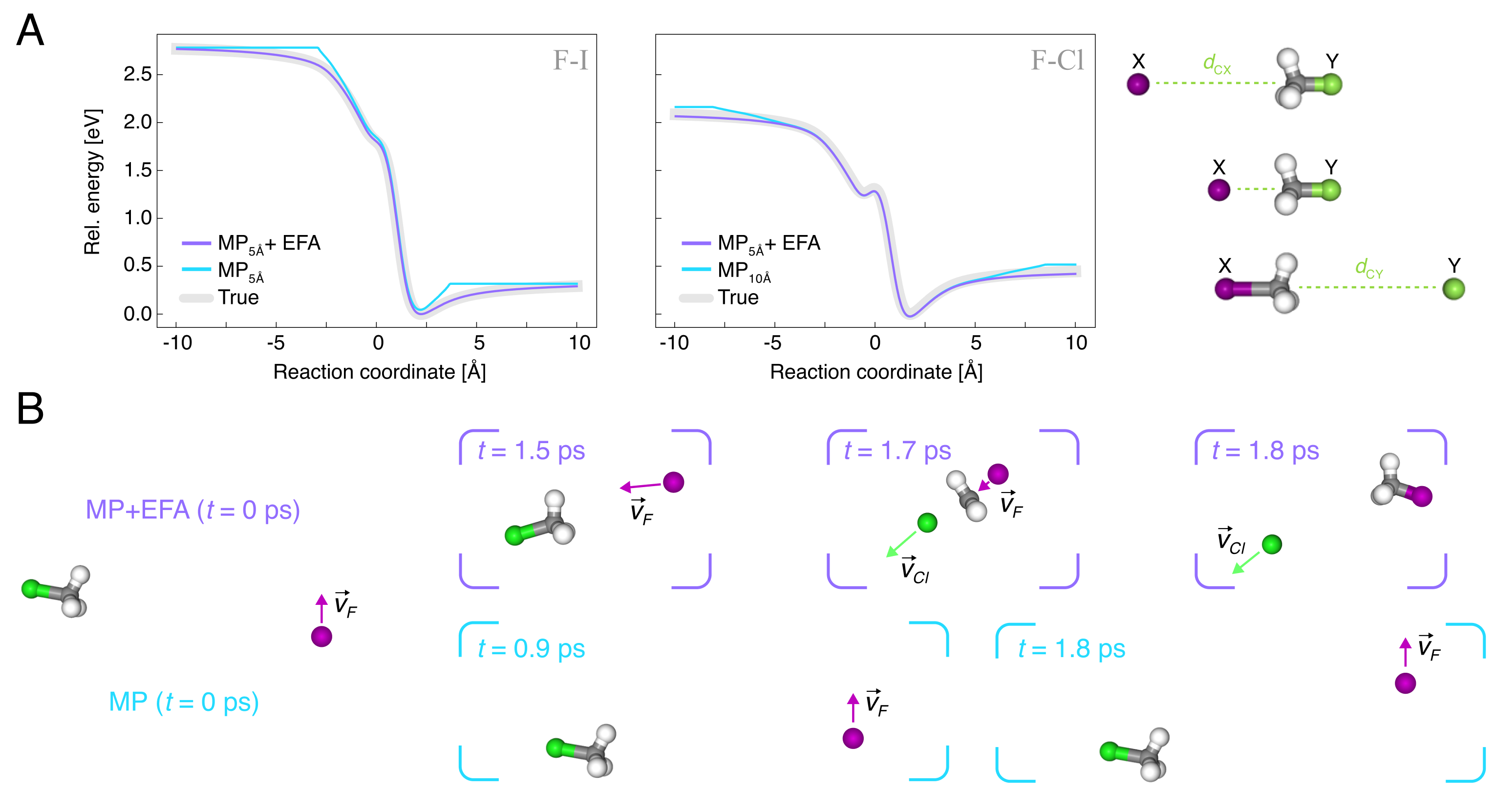}
    \begin{subfigure}{\linewidth}
        \phantomcaption{}
        \label{fig:results_panel_molecules:sn2}
    \end{subfigure}
    \begin{subfigure}{\linewidth}
        \phantomcaption{}
        \label{fig:results_panel_molecules:sn2-md}
    \end{subfigure}
    \caption{\textbf{Euclidean fast attention (EFA) for reactions. }\textbf{(A)} Illustration of an S$_\text{N}$2 reaction of the form X$^-$ + H$_3$C--Y $\rightarrow$ X--CH$_3$ + Y$^-$. The panels show a one-dimensional cut of the potential energy surface (PES) along the reaction coordinate ($d_{\text{CY}}-d_{\text{CX}}$) for two systems (X=F, Y=I and X=F, Y=Cl). A model using standard message passing (MP) with a cutoff of $r_\text{cut} = 5\,\an$ shows qualitatively wrong asymptotic behaviour (left), which also cannot be fixed by increasing the cutoff to $r_\text{cut} = 10\,\an$ (right). In contrast, models augmented with EFA reproduce the correct energy profile for all particle separations. \textbf{(B)} Snapshots at different times $t$ of two MD trajectories starting from identical initial conditions simulated with the MP+EFA model (top) and MP model (bottom), respectively. The velocities of fluorine ($\vec{v}_\text{F}$) and chlorine ($\vec{v}_\text{Cl}$) atoms are highlighted by arrows to indicate their motion. Since the MP model predicts no forces between ion and molecule beyond the cutoff, the reactants fly past each other undisturbed and no reaction occurs. In contrast, for MP+EFA, the reactants are attracted towards each other and the methyl halide molecule is being re-oriented and properly positioned for backside attack of the ion, facilitating the reaction.}
    \label{fig:sn2-reactions}
\end{figure*}

\paragraph{\texorpdfstring{S$_\text{N}$2}{SN2} Reactions} As an example of chemical reactions, we focus on prototypical S$_\text{N}$2 reactions of the type X$^-$ + H$_3$C--Y $\rightarrow$ X--CH$_3$ + Y$^-$, where X and Y are halogens (F, Cl, Br, I).\cite{unke2019physnet} We find that adding an EFA block to a local MPNN architecture ($r_{\text{cut}} = 5\,\an$) yields a $34\times$ and $8\times$ reduction in mean absolute errors (MAEs) for energy and force predictions, respectively (\autoref{tab:sn2}). \rebuttal{\label{rebuttal:sn2-add-dispersion}Merely increasing the local cutoff to $r_{\text{cut}} = 10\,\an$ or including dispersion corrections does not yield satisfactory improvements (see \autoref{si:fig:sn2-with-dispersion} and \autoref{tab:sn2}).} The reason for the substantially lower MAEs when using EFA becomes apparent when visualizing one-dimensional cuts of the potential energy surface along the reaction coordinate (\autoref{fig:results_panel_molecules:sn2}). Both models without EFA predict the wrong asymptotic behaviour whereas the model with EFA accurately describes the energy profile for the full range of particle separations. The unphysical artefacts in the long-range region of the potential energy surface predicted by the standard MP model also lead to qualitatively incorrect dynamic behaviour, i.e., only the trajectory driven by the MP+EFA model leads to a reaction resulting in CH$_3$F~+~Cl$^-$ (\autoref{fig:results_panel_molecules:sn2-md}).

\begin{figure*}[t!]
    \centering
    \includegraphics[width=\linewidth]{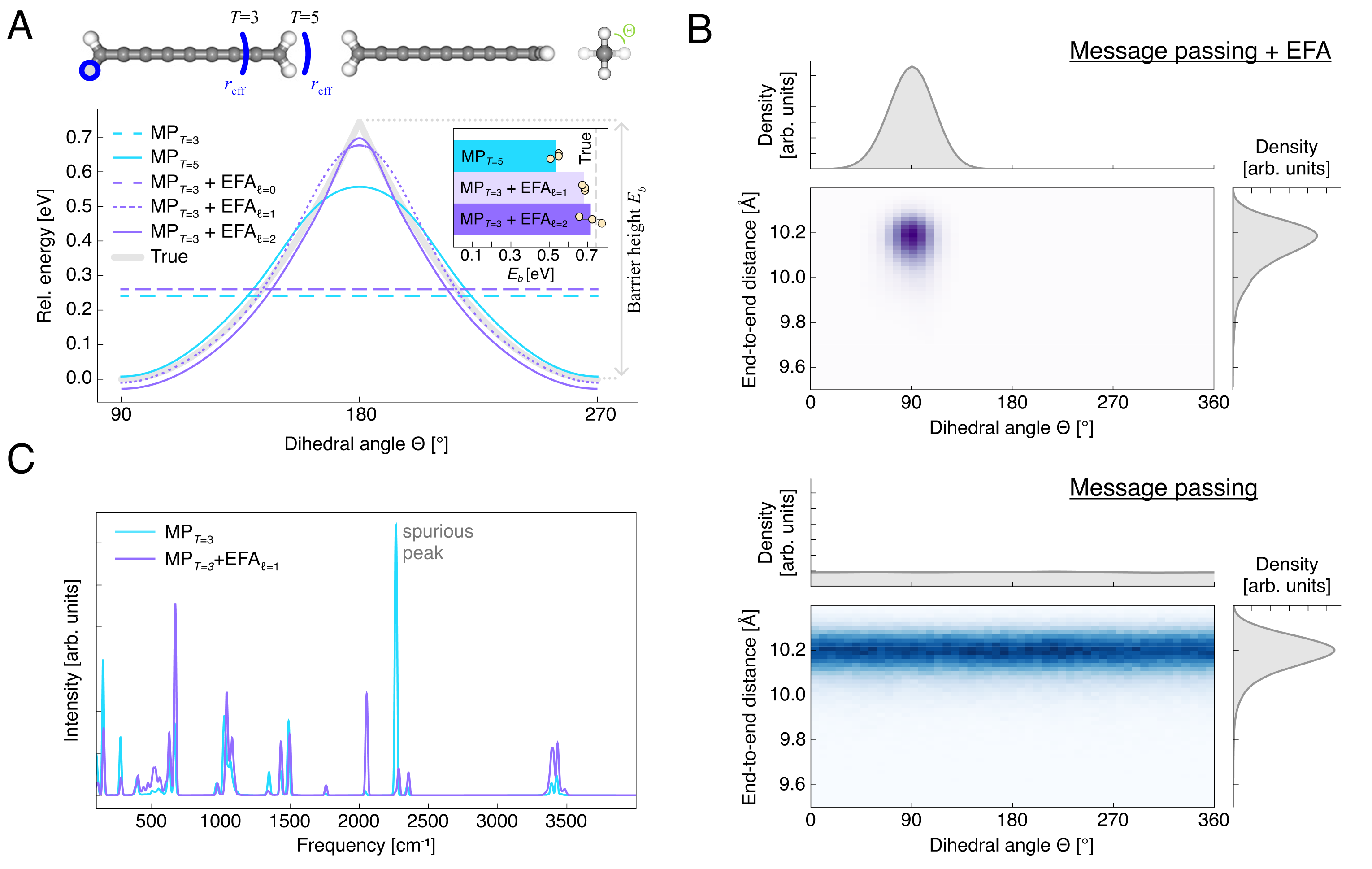}
    \begin{subfigure}{\linewidth}
        \phantomcaption{}
        \label{fig:results_panel_molecules:cumulene}
    \end{subfigure}
    \begin{subfigure}{\linewidth}
        \phantomcaption{}
        \label{fig:results_panel_cumulene:dihedral-distribution}
    \end{subfigure}
    \begin{subfigure}{\linewidth}
        \phantomcaption{}
        \label{fig:results_panel_cumulene:power-spectrum}
    \end{subfigure}
    \caption{\textbf{Euclidean fast attention (EFA) for electronically de-localised effects. }\textbf{(A)} Cumulene molecule C$_{k+2}$H$_4$ with chain length $k=7$ (the effective cutoff $r_{\text{eff}}$ from the atom marked in blue is indicated for different numbers of MP layers $T$). The panel shows the energy profile as function of the dihedral angle $\Theta$ between the terminal CH$_2$ rotors for MP ($T=3,5$) and MP+EFA ($T=3$) with different maximal degrees $\ell = 0,1,2$. The inset shows the predicted energy barrier height (only for models that do not predict a flat energy profile), yellow dots corresponds to models trained with different random seeds. \textbf{(B)} Visualisation of the configurational space spanned by the end-to-end distance (measured as the distance between the outer carbons) and the dihedral angle between the CH$_2$ rotors visited during 2\,ns molecular dynamics (MD) simulations at 300\,K in the NVT ensemble. Due to the large energy barrier in the ground truth potential energy surface at a dihedral angle of 180$^\circ$, physically accurate trajectories starting from the minimum at 90$^\circ$ are expected to only deviate from this value slightly during simulations (there is not enough energy to cross the barrier at a temperature of 300~K). The MP+EFA model (top) follows this expectation and the dihedral angle stays around 90$^\circ$, whereas the MP model (bottom) predicts wrong dynamic behaviour (all dihedral angles are sampled equally likely, consistent with the prediction of a flat energy profile). \textbf{(C)} Predicting the wrong dynamic behaviour has direct consequences on experimental observables: For example, the power spectrum extracted from the dynamics driven by the local MP model exhibits a strong spurious peak at a frequency of  $\sim$2300\,cm$^{-1}$.}
    \label{fig:results_panel_cumulene}
\end{figure*}
\paragraph{Electronic delocalisation}
So far, our analyses have focused primarily on long-range interactions that can be described as functions of atomic (or molecular) distances $r$, usually decaying proportionally to some power law $r^{-b}$. However, some chemical systems exhibit \emph{non-local} effects, which are more complicated in nature and cannot easily be described as a function of distance. Instead, they are usually characterised by a strong dependence of the energy on the relative orientation of substructures in a molecule. Here we consider the non-local effects in cumulene molecules, which have been found to be particularly challenging for both, global and local MLFFs.\cite{unke2021machine,frank2022so3krates} 

In cumulenes, the energy strongly depends on the dihedral angle $\Theta$ between the hydrogen rotors at the ends of a chain of carbon atoms (see \autoref{fig:results_panel_molecules:cumulene}). We compare a $T=3$ layer MPNN augmented with three EFA versions to regular MPNNs with $T=3$ and $T=5$ layers. EFA$_{\ell = 0}$ uses only invariant features, whereas EFA$_{\ell = 1}$ and EFA$_{\ell = 2}$ use equivariant representations up to degree $\ell = 1$ and $\ell = 2$. The MP+EFA$_{\ell = 0}$ model cannot describe the energy profile, because it only leverages pairwise distances, which cannot resolve the change in the dihedral angle sufficiently well.\cite{frank2022so3krates} As soon as equivariant representations are allowed to enter the EFA update ($\ell>0$), the energy is modelled faithfully (higher degrees enable more accurate energy barrier predictions). MP models without EFA fail and predict a flat energy profile as soon as the distance between hydrogen rotors exceeds the effective cutoff. Even when the effective cutoff is large enough ($T = 5$) the barrier height is still underestimated (inset \autoref{fig:results_panel_molecules:cumulene}), indicating over-squashing of information.\cite{alon2021on} Modelling the energy barrier correctly is crucially important for predicting the correct dynamical behaviour during MD simulations. For example, models that predict a flat energy profile incorrectly sample all possible dihedral angles between the terminal CH$_2$ rotors equally likely (\autoref{fig:results_panel_cumulene:dihedral-distribution}), which leads to wrong predictions of physical observables (\autoref{fig:results_panel_cumulene:power-spectrum}). We remark that all tested models are unable to reproduce the sharp cusp at a dihedral angle of 180$^\circ$, which is an artefact of the reference ab-inito method (see SI \nameref{si:sec:cusp-in-cumulene}).

\begin{figure*}
    \centering
    \includegraphics[width=\linewidth]{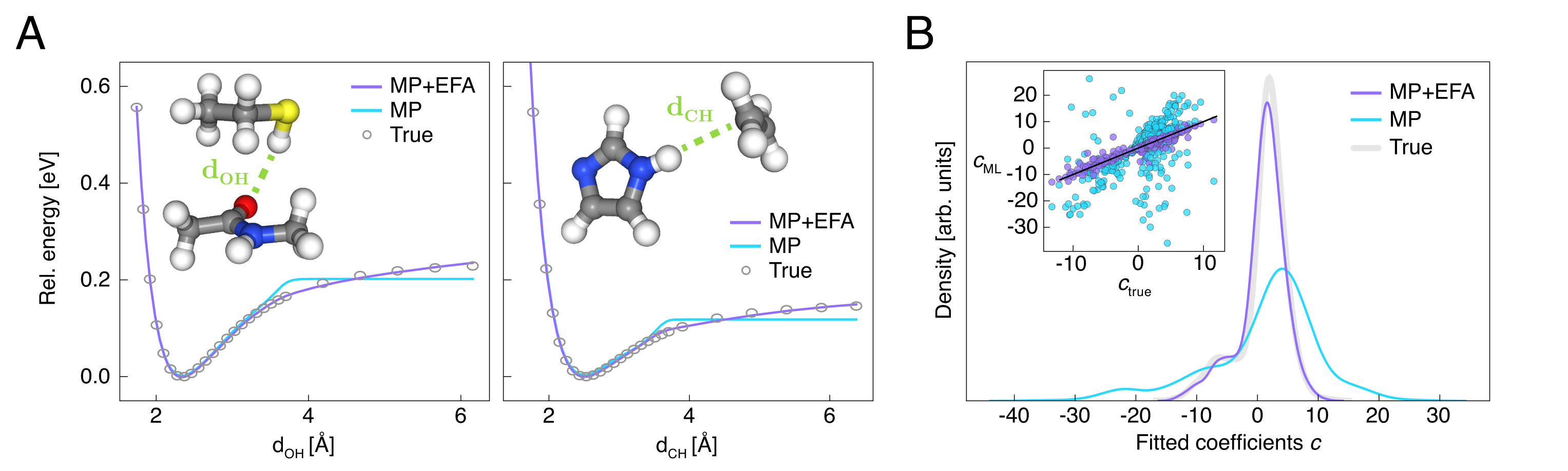}
        \begin{subfigure}{\linewidth}
        \phantomcaption{}
        \label{fig:results_panel_molecules:dimers}
    \end{subfigure}
    \begin{subfigure}{\linewidth}
        \phantomcaption{}
        \label{fig:results_panel_molecules:coefficients}
    \end{subfigure}
    \caption{\textbf{Dimers. }\textbf{(A)} Binding curves for different dimers from the DES370K data set\cite{donchev2021quantum} for a standard MP model and for a model augmented with EFA (ground truth reference values are shown with circles). 
    \textbf{(B)} Correlation (inset) and distribution of coefficients $c_1,\dots,c_6$ for long-range dimer interactions (see \autoref{eq:lr-expansion}). The Pearson correlation \wrt the coefficients fitted to the ground truth is $s=0.56$ for MP and $s=0.95$ for MP+EFA. Coefficient distributions show a similar trend, indicating an improved agreement to the ground truth for MP+EFA compared to MP.}
    \label{fig:results-panel-dimers}
\end{figure*}
\paragraph{Dimers}
Finally, we employ EFA for non-covalent interactions of dimer systems sampled from the challenging DES370K benchmark.\cite{donchev2021quantum} To perform well on this benchmark, ML models need to generalise across the functional form of different long-range interactions (e.g., electrostatics, induction, dispersion, or mixtures thereof) and molecular structures.
Consistent with our previous findings, MP fails to model dimer interaction energies as soon as the separation between the individual molecules exceeds the cutoff, whereas MP+EFA captures the energy profile faithfully (\autoref{fig:results_panel_molecules:dimers}). We also find robust generalisation of the MP+EFA model to four completely unseen dimers (\autoref{si:fig:dimer:potential}).

To further quantify the agreement of different models \wrt to the ground truth reference, we fit the long-range behaviour of the inter-molecular interactions to the functional form
\begin{align}
    V_\text{lr}(r_{AB}) = \sum_{i=1}^6 c_i \, r_{AB}^{-i},\label{eq:lr-expansion}
\end{align}
where $c_i$ are expansion coefficients and $r_{AB}$ is the distance between molecules $A$ and $B$. \autoref{eq:lr-expansion} can capture the behaviour of many physically relevant interaction types, e.g., charge-charge ($r_{AB}^{-1}$), charge-dipole ($r_{AB}^{-2}$), or dispersion ($r_{AB}^{-6}$). We calculate the Pearson correlation $s$ of coefficients for model predictions with those for the ground truth and find $s=0.56$ for the standard MP model vs.\ $s=0.95$ for the MP+EFA model. Further, we find excellent agreement between the coefficient distribution for the ground truth and the MP+EFA model, whereas the MP model yields a qualitatively wrong distribution (\autoref{fig:results_panel_molecules:coefficients} and \autoref{si:fig:dimer:distribution} for individual $c_i$).

\rebuttal{
\paragraph{Systems without strong long-range effects}
For completeness, we evaluate standard MPNNs and EFA-augmented versions on the materials investigated in Ref.~\citenum{sauceda2022bigdml}, which are dominated by local interactions.\label{rebuttal:materials_benchmarks} As expected, since these systems exhibit no strong long-range effects, both MP and MP+EFA models perform comparably (\autoref{tab:bigdml}). Similar results can be observed on standard molecular benchmarks where local interactions dominate (see SI \nameref{si:sec:established-local-benchmarks}).\label{rebuttal:standard-benchmarks}}

\begin{table}[ht!]
\centering
{ 
\setlength{\tabcolsep}{4pt} 
\begin{tabular}{llccc}
\toprule
\textbf{System} & \textbf{Metric} & \textbf{BIGDML} & \textbf{MP} & \textbf{MP+EFA} \\ 
\midrule
\multirow{2}{*}{Graphene} & \textit{Energy} & 0.04 & 0.02 & 0.02 \\
 & \textit{Forces} & 7.1 & 4.1 & 4.4 \\
\midrule 
\multirow{2}{*}{Na} & \textit{Energy} & 0.3 & 1.0 & 0.8 \\
 & \textit{Forces} & 1.2 & 13.1 & 13.2 \\
\midrule
\multirow{2}{*}{Pd32H} & \textit{Energy} & 0.3 & 0.2 & 0.2 \\
 & \textit{Forces} & 17.9 & 10.7 & 9.8 \\
\midrule
\multirow{2}{*}{Pd$_1$MgO} & \textit{Energy} & 0.9 & 0.6 & 0.6 \\
 & \textit{Forces} & 41.2 & 27.7 & 27.8 \\
\midrule
\multirow{2}{*}{Pd\,1000K} & \textit{Energy} & 0.3 & 0.3 & 0.8 \\
 & \textit{Forces} & 21.2 & 16.2 & 16.8 \\
\midrule
\multirow{2}{*}{Pd\,\text{500K}} & \textit{Energy} & 0.1 & 0.2 & 0.1 \\
 & \textit{Forces} & 7.1 & 5.9 & 6.4 \\
\bottomrule
\end{tabular}
}
\caption{\rebuttal{\textbf{Periodic systems without strong long-range effects.} Mean absolute error (MAE) for energy (meV/atom) and forces (meV/$\si{\angstrom}$) for the systems from Ref.~\citenum{sauceda2022bigdml}. We also compare to BIGDML,\cite{sauceda2022bigdml} a kernel method specifically designed for the treatment of periodic systems. Results averaged over two runs.}}
\label{tab:bigdml}
\end{table}

\phantomsection
\section{Discussion and Conclusion}
\label{sec:discussion-and-conclusion}

For computational efficiency, many machine learning force fields (MLFFs) use a local cutoff, which (by design) prevents modelling global and long-range interactions. While global interactions are often weaker than short-range interactions, they are crucial for accurately modelling large systems like proteins and solids. \cite{stohr2019theory} Developing methods to faithfully treat these global effects is therefore essential for scaling MLFFs to larger, chemically relevant systems.

Different approaches have been proposed to overcome the limitations of strictly local MLFFs. They can be broadly categorised into two classes: Methods that (i) augment otherwise local models with global correction terms or (ii) abandon the concept of strict locality altogether by instead learning global representations. The first category typically relies on a physically motivated description of specific long-range effects.\cite{muhli2021machine, westermayr2022long, artrith2011high, morawietz2012neural, ko2021fourth, unke2019physnet, grisafi2019incorporating, unke2021spookynet, pagotto2022predicting, li2023long, loche2024fast} These approaches usually involve learning intermediate, local quantities, which are then used as parameters in fixed interaction terms. \rebuttal{
The computational complexity of naively evaluating these terms typically scales quadratically with the number of atoms, but can be sped up with techniques like the fast multipole method,\cite{rokhlin1985rapid} which allows their application to large systems.\label{rebuttal:fast_multipole_method} A popular example for this kind of approach is to predict partial charges, either directly,\cite{unke2019physnet, unke2021spookynet, artrith2011high, morawietz2012neural} or via atomic electronegativities, which are used as inputs to a charge equilibration scheme (to account for non-local charge transfer effects).\cite{ko2021fourth} The obtained charges can then be used to calculate electrostatic interactions via Coulomb's law or (for periodic systems) with Ewald summation.\cite{ewald1921berechnung} \label{rebuttal:latent_ewald_methods} Recent works even propose to learn ``latent charges'', which may not necessarily have a physical meaning, but can be used to evaluate Ewald-like terms to model long-range interactions in periodic systems (see SI \nameref{si:sec:relation-to-ewald-mp}).\cite{kim2024learning,cheng2025latent}}

Most of these methods define fixed functional forms, implicitly assuming that all relevant long-range interactions can be modelled this way. While this may be enough for some systems with known types of interactions, recent studies highlight that complex systems, such as proteins, can display long-range information exchange with unique and non-trivial interaction strengths.\cite{gori2023second} In contrast, approaches in the second category instead aim to learn the correct interaction patterns directly from data without relying on domain knowledge or making assumptions about the nature of the underlying interactions. \rebuttal{\label{rebuttal:rephrasing_limitations}However, achieving this without sacrificing computational efficiency is difficult and existing approaches are typically tailored to very specific systems and interaction types\cite{unke2021spookynet, frank2022so3krates, batatia2023equivariant} or assume a fixed reference frame\cite{kosmala2023ewald, wang2024enhancing} to make the implementation more straightforward.}

To address this challenge, our work proposes Euclidean fast attention (EFA), which allows to learn global atom/node representations for molecules (and other data embedded in Euclidean space) with linear time and memory complexity and respects all relevant physical symmetries. A core component of EFA are novel Euclidean rotary positional (ERoPE) encodings, which are combined with a linear-scaling attention-like mechanism and integration over the unit sphere. We show that standard MPNNs have systematic shortcomings in capturing the global correlations essential for chemical accuracy. Augmenting them with EFA resolves these issues, enabling the learning of global interactions. We demonstrate EFA's broad utility across diverse systems, including reactions, electronic delocalisation, and dimers. 

A natural question to ask is whether it is possible to build accurate models purely out of EFA blocks. We believe that further algorithmic improvements would be necessary to make this feasible. Since EFA relies on a numerical solution of the surface integral over $S^2$ via Lebedev quadrature,\cite{lebedev1976quadratures} it is best suited to model ``low frequency'' functions that vary slowly with distance (and for which small integration grids are sufficiently accurate). Our results demonstrate that this seems to be no fundamental issue for modelling non-local and long-ranged interactions, but likely causes problems when describing strong short-ranged interactions (when EFA is not combined with a local model), which can vary rapidly for small changes in distance. Future work should thus focus on augmenting existing local architectures with EFA to increase their accuracy for long-ranged interactions and on developing faster numerical integration methods for the special structure of the surface integral in \autoref{eq:euclidean-fast-attention-equivariant}.

\phantomsection
\section{Methods} 
\label{sec:methods}

\subsection{Euclidean Fast attention (EFA)}
In the following, we only describe the implementation of EFA in its most general form (\autoref{eq:euclidean-fast-attention-equivariant}), which assumes equivariant features (the mechanism for invariant features in \autoref{eq:euclidean-fast-attention} can be recovered as a special case). We first describe all individual components and then show how they are combined to the full EFA mechanism.

\paragraph{Equivariant features} On a high level, equivariant features can be thought of as containing additional ``directional information''. Under transformations of the coordinate system, the numerical values of equivariant features may change in a complicated manner, but they still encode ``the same'' directional information (but transformed accordingly). Here, we briefly describe a particular type of equivariant features introduced in Ref.~\citenum{unke2024e3x}. 

The equivariant features we consider consist of irreducible representations (irreps) of the orthogonal group in three dimensions $\mathrm{O}(3)$ (rotations and reflections). Features of degree $\ell$ are denoted as $\boldsymbol{x}^{(\ell)} \in \mathbb{R}^{P \times (2\ell + 1) \times H}$, where $P$ is the size of the ``parity axis'' (either $1$ or $2$) and $H$ is the feature space dimension. The parity of an irrep can be either even ($+1$) or odd ($-1$) and determines how it behaves under reflections (it either changes sign or not), whereas the degree $\ell$ determines the behaviour under rotations (an irrep of degree $\ell$ has $2\ell+1$ components/is described by $2\ell+1$ numbers). For example, irreps with $\ell=0$ are described by a single number and do not change (are invariant) under rotations, whereas irreps with $\ell=1$ consist of three components and rotate similar to ordinary three-dimensional vectors. When irreps have parity $+1$ for even $\ell$ and $-1$ for odd $\ell$, we refer to them as ``tensors'', and when they have the opposite parity, we call them ``pseudotensors''. Features either consist only of tensors (in which case $P = 1$), or of both tensors and pseudotensors (in which case $P = 2$ and irreps of even/odd parity are stored in separate slices, i.e., indices $0$ and $1$, of the parity axis). For a more detailed overview, we refer the reader to Ref.~\citenum{unke2024e3x}.

We are usually working with the concatenation of features starting from the lowest degree $\ell = 0$ up to some maximum degree $L$, which we denote as $\boldsymbol{x} \in \mathbb{R}^{P \times (L + 1)^2 \times H}$ (each degree $\ell$ contributes a slice of size $2\ell + 1$, giving a total size $(L + 1)^2$ for the ``degree axis''). We use $\boldsymbol{x}_m$ to refer to the representation of the $m$-th atom and $\boldsymbol{x}^{(\ell_{\pm})}$ to refer to the ``slice'' of irreps of degree $\ell$ with parity $p = \pm 1$. Invariant features, for example those used in SchNet\cite{schutt2018schnet} or PhysNet,\cite{unke2019physnet} can be considered as a special case with $L = 0$ and $P=1$, i.e., they only consist of irreps with degree $0$ and even parity.

Two irrep representations $\boldsymbol{x}$ and $\boldsymbol{y}$ can be ``coupled'' via tensor product contractions\cite{unke2021se} to produce new features $\boldsymbol{z}$. The irreps of degree $c$ and parity $\gamma$ of the new features $\boldsymbol{z}$ are given by
\begin{align}
    \boldsymbol{z}^{(c_\gamma)} = \sum_{(a_\alpha, b_\beta)} \boldsymbol{x}^{(a_\alpha)} \otimes^{(c_\gamma)} \boldsymbol{y}^{(b_\beta)}, 
    \label{eq:tensor-product-contraction}
\end{align}
where the sum runs over all combinations of irreps with degrees $a + b = c$ and parities $\alpha \cdot \beta = \gamma$ of the input features $\boldsymbol{x}$ and $\boldsymbol{y}$. Evaluating the `$ \otimes^{(c_\gamma)}$' operation in \autoref{eq:tensor-product-contraction} involves a summation over components of the tensor (outer) product weighed with so-called Clebsch-Gordan coefficients (see Ref.~\citenum{unke2024e3x} for implementation details). 

Performing all possible tensor product contractions between irreps of maximal degree $L_x$ and $L_y$ up to a maximal output degree $L_z \leq L_x + L_y$ is written as 
\begin{equation}
    \boldsymbol{z} = \boldsymbol{x} \Motimes_{L_z}^{L_x\,L_y} \boldsymbol{y},
\label{eq:concatenated-tensor-product-contraction}
\end{equation}
such that $\boldsymbol{z} \in \mathbb{R}^{P \times (L_z+1)^2 \times H}$. In other words, this is just a compact notation for the concatenation of all tensor product contractions (\autoref{eq:tensor-product-contraction}) for $c = 0, \dots, L_z$ and $\gamma = \pm 1$.

\paragraph{Euclidean rotary positional encodings (ERoPE)}
Following typical implementations of the RoPE mechanism,\cite{su2024roformer} ERoPE is implemented in a slightly different (but mathematically equivalent) manner than what is suggested in \autoref{eq:euclidean-rotary-positional-encoding}. Consider a complex number $c=a+bi$  ($c\in\mathbb{C}$ and $a,b\in\mathbb{R}$). Recall that, to encode a position $\vec{r}\in\mathbb{R}^3$ in Euclidean space into $c$ with ERoPE, we calculate
\begin{equation*}
\mathrm{ERoPE}_{\vec{u}}(c,\vec{r}) = c \cdot e^{i\omega\vec{u}\cdot\vec{r}}\,.
\end{equation*}
To avoid computations with complex numbers, we can instead collect the real and imaginary parts of $c$ into a vector $\boldsymbol{x}=[a\ b]^\T \in \mathbb{R}^2$ and perform an equivalent encoding of $\vec{r}$ into $\boldsymbol{x}$ as
\begin{equation*}
\mathrm{ERoPE}_{\vec{u}}(\boldsymbol{x},\vec{r}) = \boldsymbol{M}\boldsymbol{x}\,,
\end{equation*}
where $\boldsymbol{M}$ is the $2\times 2$ rotation matrix
\begin{equation*}
\begin{bmatrix}
\cos(\omega\vec{u}\cdot\vec{r}) & -\sin(\omega\vec{u}\cdot\vec{r})\\
\sin(\omega\vec{u}\cdot\vec{r}) & \phantom{-}\cos(\omega\vec{u}\cdot\vec{r})
\end{bmatrix}\,.
\end{equation*}
This formulation can be extended to higher-dimensional vectors $\boldsymbol{x}\in\mathbb{R}^{H}$ with $H=2K$, where $\boldsymbol{M}$ is now a $(2K)\times (2K)$ block-diagonal matrix consisting of $2\times2$ rotation matrices and we allow different coefficients $\omega_k$ for each of the $K$ rotations matrices (the motivation for this is explained below). For high-dimensional vectors $\boldsymbol{x}$, instead of evaluating $\boldsymbol{M}\boldsymbol{x}$ via matrix multiplication, it is more efficient (due to the sparsity of $\boldsymbol{M}$) to compute
\begin{align}
    \mathrm{ERoPE}_{\vec{u}}(\boldsymbol{x}, \vec{r}) &=
    \begin{bmatrix}
    \cos(\omega_1 \vec{u} \cdot \vec{r})\\
    \cos(\omega_1 \vec{u} \cdot \vec{r})\\
    \vdots\\
    \cos(\omega_K \vec{u} \cdot \vec{r})\\
    \cos(\omega_K \vec{u} \cdot \vec{r})
    \end{bmatrix}
    \odot
    \begin{bmatrix}
    x_1\\
    x_2\\
    \vdots\\
    x_{2K-1}\\
    x_{2K}
    \end{bmatrix} \nonumber\\ 
    &\,\,\,\,\,\,\,\,+ \, 
    \begin{bmatrix}
    \sin(\omega_1 \vec{u} \cdot \vec{r})\\
    \sin(\omega_1 \vec{u} \cdot \vec{r})\\
    \vdots\\
    \sin(\omega_K \vec{u} \cdot \vec{r})\\
    \sin(\omega_K \vec{u} \cdot \vec{r})
    \end{bmatrix}
    \odot
    \begin{bmatrix}
    -x_2\\
    x_1\\
    \vdots\\
    -x_{2K}\\
    x_{2K-1}
    \end{bmatrix}, 
\label{eq:erope_actual_implementation}
\end{align}
where `$\odot$' denotes element-wise multiplication and $x_j$ is the $j$-th entry of $\boldsymbol{x}$. For vectors with an odd number of dimensions, i.e., $H=2K-1$, the missing entry $x_{2K}$ can be simply replaced by zero ($\boldsymbol{x}$ is zero-padded to an even number of dimensions). 

Expressed in words, if we interpret a vector $\boldsymbol{x}\in\mathbb{R}^{2K}$ as consisting of the concatenated real and imaginary parts of the (complex) entries of a vector $\tilde{\boldsymbol{x}}\in\mathbb{C}^{K}$, then applying \autoref{eq:erope_actual_implementation} to $\boldsymbol{x}$ (with $\omega_k=\omega$) is equivalent to applying \autoref{eq:euclidean-rotary-positional-encoding} to $\tilde{\boldsymbol{x}}$. The motivation for allowing different $\omega_k$ for each entry $k$ is that it increases the expressivity of the operation: Recall that after integration over $S^2$, encoding positions $\vec{r}_m$ and $\vec{r}_n$ into vectors $\boldsymbol{q}_m$ and $\boldsymbol{k}_n$ with ERoPE and computing their scalar product essentially corresponds to scaling $\langle \boldsymbol{q}_m, \boldsymbol{k}_n\rangle$ with $\mathrm{sinc}(\omega r_{mn})$ (see \autoref{eq:surface-integral-solution} and SI \nameref{si:sec:surface-integral}). When different $\omega_k$ are used for each entry, instead of applying a uniform scaling factor, the contribution of different entries to the scalar product can be scaled independently by factors of $\mathrm{sinc}(\omega_k r_{mn})$, which allows to express a more complicated radial dependence. 

The extension of ERoPE to equivariant features $\boldsymbol{x} \in \mathbb{R}^{P \times (L + 1)^2 \times H}$ is straightforward: We simply consider $\boldsymbol{x}$ as a collection of $P \times (L + 1)^2$ vectors of dimension $H$, to which \autoref{eq:erope_actual_implementation} is applied separately. In other words, ERoPE is applied along the ``feature axis'' (of size $H$), whereas the ``parity axis'' (of size $P$) and the ``degree axis'' (of size $(L + 1)^2$) are treated as ``batch dimensions''.

\paragraph{Spherical Harmonics} The spherical harmonics are functions defined on the surface of the unit sphere $S^2 \coloneqq \{\vec{u} \in \mathbb{R}^3: \lVert \vec{u} \rVert = 1\}$. We follow the conventions used in Ref.~\citenum{unke2024e3x} and define them as real and vector-valued, i.e., $\boldsymbol{Y}_{\ell}: S^2 \mapsto \mathbb{R}^{2\ell+1}$. For example, the first two of these functions (using the Racah normalisation convention) are given by $\boldsymbol{Y}_{0}(\vec{u}) = 1$ and $\boldsymbol{Y}_{1}(\vec{u}) = \vec{u}$. In general, the individual components of $\boldsymbol{Y}_{\ell}$ are polynomials of degree $\ell$ in the $x$,$y$, and $z$ components of the vector $\vec{u}$ (see Ref.~\citenum{unke2024e3x} for a general definition for arbitrary degrees~$\ell$).

\paragraph{Lebedev quadrature} A Lebedev quadrature\cite{lebedev1976quadratures} is a numerical approximation to the surface integral of a function $f$ over the unit sphere $S^2$
\begin{equation}
\int_{S^2} f(\vec{u})\ \mathrm{d}\vec{u} \approx 4\pi \sum_{j=1}^{\numgridpoints} \lambda_j f(\vec{u}_j)\,,
\label{eq:lebedev-quadrature}
\end{equation}
where $\vec{u}_j \in S^2$ are grid points and $\lambda_j\in \mathbb{R}$ the corresponding quadrature weights. The grid can be made denser, i.e., the number of grid points $\numgridpoints$ can be increased, to make the approximation more accurate. Similar to one-dimensional Gaussian quadratures, Lebedev quadratures have the important property that integration of polynomials up to a certain degree is \emph{exact}. For example, a grid with just $\numgridpoints=6$ points integrates polynomials of up to degree $3$ (such as the spherical harmonics with $\ell \leq 3$) without error.\cite{lebedev1976quadratures}

\paragraph{Implementation of EFA}
Given atomic features $\boldsymbol{x}_{m}^{(\ell_p)} \in \mathbb{R}^{P\times 2\ell + 1 \times H}$, queries, keys and values are calculated as
\begin{equation*}
 \begin{aligned}
    \boldsymbol{q}_{m}^{(\ell_p)} &= \boldsymbol{x}_{m}^{(\ell_p)} W^{(\ell_p)}_q\,,\\
    \boldsymbol{k}_{m}^{(\ell_p)} &= \boldsymbol{x}_{m}^{(\ell_p)} W^{(\ell_p)}_k\,,\\
    \text{and}\quad \boldsymbol{v}_{m}^{(\ell_p)} &= \boldsymbol{x}_{m}^{(\ell_p)} W^{(\ell_p)}_v\,.
\end{aligned}  
\end{equation*}
Here, $W_{q}^{(\ell_p)}\in \mathbb{R}^{H \times D_{qk} }$, $W_k^{(\ell_p)} \in \mathbb{R}^{H \times D_{qk}}$ and $W_v^{(\ell_p)} \in \mathbb{R}^{H \times D_v}$ are separate weight matrices for each combination of degree $\ell$ and parity $p$ and matrix multiplication is performed along the ``feature axis'', e.g., $\boldsymbol{x}^{(\ell_p)}W^{(\ell_p)}_v \in \mathbb{R}^{P\times 2\ell + 1 \times D_v}$. Atomic positions are encoded into query and key vectors with ERoPE
\begin{align}
    \tilde{\boldsymbol{q}}_{m, \vec{u}}^{(\ell_p)} \coloneqq \text{ERoPE}_{\vec{u}}\bigg(\vec{r}_m, \boldsymbol{\psi}\Big(\boldsymbol{q}^{(\ell_p)}_m\Big)\bigg),\nonumber
    \\
    \tilde{\boldsymbol{k}}_{m, \vec{u}}^{(\ell_p)} \coloneqq \text{ERoPE}_{\vec{u}}\bigg(\vec{r}_m, \boldsymbol{\psi}\Big(\boldsymbol{k}_m^{(\ell_p)}\Big)\bigg),\nonumber
\end{align}
where $\boldsymbol{\psi}$ can be any feature map that preserves equivariance (see \autoref{eq:similarity_kernel_rewrite}). Here we choose a simple identity function or gated\cite{weiler20183d} GELU\cite{hendrycks2016gaussian} non-linearity, which is applied element-wise along the ``feature axis'' of size $D_{qk}$.

The individual $\tilde{\boldsymbol{q}}_{m, \vec{u}}^{(\ell_p)}$, $\tilde{\boldsymbol{k}}_{m, \vec{u}}^{(\ell_p)}$, and $\boldsymbol{v}_m^{(\ell_p)}$ for all of the $N$ atoms are concatenated row-wise to give
\begin{equation*}
    \begin{aligned}
    \tilde{\boldsymbol{Q}}_{\vec{u}} &\in \mathbb{R}^{N \times P_{qk} \times (L_{qk}+1)^2 \times D_{qk}}\,,\\
    \tilde{\boldsymbol{K}}_{\vec{u}} &\in \mathbb{R}^{N \times P_{qk} \times (L_{qk} + 1)^2 \times D_{qk}}\,,\\
    \boldsymbol{V} &\in \mathbb{R}^{N \times P_{v} \times (L_{v} + 1)^2 \times D_{v}}\,.
\end{aligned}
\end{equation*}
Note that queries $\tilde{\boldsymbol{Q}}_{\vec{u}}$ and keys $ \tilde{\boldsymbol{K}}_{\vec{u}}$ must have the same maximal degree $L_{qk}$, parity size $P_{qk}$, and feature dimension $D_{qk}$, whereas the corresponding choices for the values $\boldsymbol{V}$ may be different. Additionally, we include $\vec{u}$ as a subscript for $\tilde{\boldsymbol{Q}}_{\vec{u}}$ and $\tilde{\boldsymbol{K}}_{\vec{u}}$ to make explicit that their entries depend on the choice of $\vec{u}$ used for the ERoPE encoding of the atomic positions.

Using slight abuse of notation (which is motivated below), we can now ``translate'' the linear-scaling attention-like mechanism introduced in \autoref{eq:linear-scaling-attention-no-denominator} to 
\begin{align}
    \boldsymbol{B}_{\vec{u}} \coloneqq  \hat{\boldsymbol{Q}}_{\vec{u}} \Big( \, \hat{\boldsymbol{K}}_{\vec{u}}^\T \, \boldsymbol{V} \, \Big), \label{eq:efa-B-tensor}
\end{align}
where $\boldsymbol{B}_{\vec{u}} \in \mathbb{R}^{N \times P_v \times (L_v + 1)^2 \times D_v}$. Here, the ``outer product'' $\hat{\boldsymbol{K}}_{\vec{u}}^\T \, \boldsymbol{V}$ of keys and values can be computed with time complexity $\mathcal{O}(N)$ and should be thought of as evaluating to an array of shape $[P_{qk} \times (L_{qk}+1)^2 \times D_{qk} \times P_v \times  (L_v + 1)^2 \times D_v]$. The subsequent multiplication with $\hat{\boldsymbol{Q}}_{\vec{u}}$ from the left side  should be understood as reducing over (element-wise multiplication followed by summation) the last three axes of $\hat{\boldsymbol{Q}}_{\vec{u}}$ and the first three axes of $\hat{\boldsymbol{K}}_{\vec{u}}^\T \, \boldsymbol{V}$. This operation also has a time complexity of $\mathcal{O}(N)$, which makes the full operation (\autoref{eq:efa-B-tensor}) scale linearly in the number of atoms. We use the suggestive ``matrix-like'' notation in \autoref{eq:efa-B-tensor}, because when $P_{qk}=P_{v}=1$ and $L_{qk}=L_{v}=0$ (special case for ``ordinary'' invariant features), many axes of the involved arrays have size $1$ and can be omitted (``squeezed'') for simplicity. In this special case, $\tilde{\boldsymbol{Q}}_{\vec{u}}, \tilde{\boldsymbol{K}}_{\vec{u}} \in \mathbb{R}^{N\times D_{qk}}$ and $V \in \mathbb{R}^{N\times D_v}$, so \autoref{eq:efa-B-tensor} is correctly described with conventional notation for matrix operations.

The full equivariant Euclidean fast attention (EFA) update (\autoref{eq:euclidean-fast-attention-equivariant}) is implemented as
\begin{equation}
    \text{EFA}(\mathcal{X}, \mathcal{R}) =\sum_{j=1}^{\numgridpoints} \lambda_j \,  \boldsymbol{B}_{\vec{u}_j}  \Motimes_{L_{\text{out}}}^{ L_v\,L_Y } \boldsymbol{Y}_{\vec{u}_j}\,,
    \label{eq:efa-update-methods}
\end{equation}
where we replace the integral over $S^2$ with a Lebedev quadrature and $\boldsymbol{Y}_{\vec{u}} \in \mathbb{R}^{1\times 1\times(L_Y + 1)^2\times1}$ is the concatenation of spherical harmonics vectors $\boldsymbol{Y}_\ell(\vec{u})$ up to maximal degree $L_Y$ with additional ``dummy axes'' of size $1$ that are broadcasted over the corresponding axes of $\boldsymbol{B}_{\vec{u}}$, so that the tensor product contractions (see \autoref{eq:concatenated-tensor-product-contraction}) are well-defined. When $L_Y = 0$ (and $\boldsymbol{B}_{\vec{u}_j}$ contains only invariant information, i.e., $P_v = 1$ and $L_v=0$), the invariant form of EFA (\autoref{eq:euclidean-fast-attention}) is recovered. 

\paragraph{Numerical accuracy} Since the integration over $S^2$ is performed numerically in \autoref{eq:efa-update-methods}, the accuracy of the rotational invariance/equivariance of the EFA update is directly related to the precision of the Lebedev quadrature. To better understand the limitations of this approach, we point out that the function that is integrated in \autoref{eq:efa-update-methods} basically corresponds to a linear combination of terms of the form $\sin(\omega_k \vec{u}\cdot\vec{r}_{mn})\cdot\mathrm{poly}_\ell(\vec{u})$ and $\cos(\omega_k \vec{u}\cdot\vec{r}_{mn})\cdot\mathrm{poly}_\ell(\vec{u})$, where $\mathrm{poly}_\ell(\vec{u})$ is some polynomial of degree $\ell \leq L_Y$ in the $x$, $y$, and $z$ components of the vector $\vec{u}$ (the sine/cosine components stem from ERoPE, whereas the polynomials come from the spherical harmonics). As mentioned earlier, polynomials up to a certain degree are integrated exactly when using Lebedev quadratures. While, the sine/cosine components are not polynomials, it is still possible to estimate a sort of ``pseudo-degree'' for them: We can approximate these terms by a Taylor series in $b\coloneqq \omega_k \vec{u}\cdot\vec{r}_{mn}$ around zero. Clearly, the larger $b$ is, the more terms are required in the Taylor series for a good approximation, which corresponds to a larger pseudo-degree. The largest ``total degree'' of the terms in the integral is then given by the sum of the highest degree $L_Y$ of the spherical harmonics and the estimated pseudo-degree of the sine/cosine terms. Assuming a fixed budget of grid points $\numgridpoints$, the integration will be sufficiently accurate up to some $b_{\text{max}}$, which yields the inequaltiy $\omega_k r_{nm} \leq b_{\text{max}}$. This can be used to define a maximal frequency 
\begin{equation}
    \omega_{\text{max}} = \frac{b_{\text{max}}}{r_{\text{max}}} \label{eq:max-frequency-bound}
\end{equation}
up to which (almost) exact rotational invariance/equivariance is ensured. Here, $r_{\text{max}}$ is the maximal expected distance between atoms (which can be estimated from the training data), and the value of $\omega_{\text{max}}$ can be chosen accordingly. Alternatively, the number of grid points $\numgridpoints$ can be increased to make $b_{\text{max}}$ larger, which in turn allows choosing a larger $\omega_{\text{max}}$. Because the analytic solution of the integral is known (see SI \nameref{si:sec:surface-integral}), the value of $b_{\text{max}}$ (for a given $\numgridpoints$) can be easily determined numerically. In this work, we define $b_{\text{max}}$ as the largest value for which the absolute deviation between numerical and analytic solution stays below a value of $10^{-5}$ (which roughly corresponds to the numerical precision of single precision floating point arithmetic). We confirm empirically that our implementation of EFA (\autoref{eq:efa-update-methods}) is rotationally invariant/equivariant up to the desired numerical precision in \autoref{si:fig:efa-vs-bessel} and provide precomputed values for $b_\text{max}$ as function of $\numgridpoints$ in \autoref{si:tab:bmax-to-lebedev-num}.

\rebuttal{
\label{rebuttal:grid_size_considerations}
It should be pointed out that the total cost of evaluating EFA scales as $\mathcal{O}(N\cdot \numgridpoints)$, i.e., it is linear in both the number of particles $N$ and the number of grid points $\numgridpoints$. Thus, a computational advantage over quadratically-scaling methods is only obtained if $\numgridpoints<N$. For this reason, $\numgridpoints$ should be kept as small as possible for maximum efficiency. For many applications, a more lenient accuracy threshold than $10^{-5}$ (and therefore smaller $\numgridpoints$ than those listed in \autoref{si:tab:bmax-to-lebedev-num}) may be acceptable. Empirically, we find that many long-range interactions can be captured accurately with low values of $\omega_{\text{max}}$, so small grids are typically feasible. For maximum performance, we recommend benchmarking different settings to determine an optimal trade-off between accuracy and efficiency ($\numgridpoints$ can even be modified after model training, as it only influences the accuracy of numerical integration, but not the learned interactions). Finally, we note that EFA only scales linearly with respect to $N$ under the assumption that $G$ can be kept constant when increasing the number of particles. In cases where an increase of system size also necessitates increasing $b_{\text{max}}$ (see \autoref{eq:max-frequency-bound}), $G$ needs to be increased accordingly to keep the same numerical accuracy.} 

\subsection{Neural Network Implementation} \label{sec:neural-network-implementation}
For the experiments performed in this work, we choose an equivariant MPNN in the spirit of tensor field networks\cite{thomas2018tensor} or NequIP.\cite{batzner20223} This base model is then optionally augmented with EFA blocks, and we compare the performance of the different architectural variants. Because our analyses do not depend on particular model details, but rather investigate general limitations inherent to all strictly local MLFFs, we expect our results to also transfer to other choices of local models (which can be augmented with EFA blocks in the same manner). MPNNs are merely chosen as a representative local model due to their popularity, and because an effective cutoff (extending beyond the local cutoff) only exists for MPNN-like models, and needs to be carefully considered during the analysis of results (other local models, such as Behler-Parrinello neural networks,\cite{behler2007generalized} can be regarded as an MPNN-like model with $T=1$ in our considerations).  

For the implementations of equivariant operations we use the E3x library\cite{unke2024e3x} which is build on top of JAX\cite{jax2018github} and FLAX.\cite{flax2020github} NumPy~\cite{harris2020array} and ASE~\cite{larsen2017atomic} are used for data pre- and post-processing.

\paragraph{Message Passing Neural Networks}
We follow the typical design of MPNNs discussed in \nameref{subsec:limitations-of-local-models} and iteratively update initial embeddings using repeated application of 
\begin{align}
    \mathcal{X}^{[t+1]} = \text{MessagePassing}\left(\mathcal{X}^{[t]}, \mathcal{R}\right)
    \label{eq:mp-update-block}
\end{align}
(the MessagePassing block is described below). The atomic representations with $\ell = 0$ and even parity are initialised via learned atom type embeddings dependent on the atomic numbers $\mathcal{Z} \coloneqq \{z_1, \dots, z_N\,|\,z_m \in \mathbb{N}_+\}$. In case of the charged cluster experiments (see \nameref{sec:idealised-systems}), we instead use the atomic charges $\mathcal{Q} \coloneqq \{q_1, \dots, q_N\,|\,q_m \in \mathbb{Z}\}$ to assign the embeddings. All other degree and parity channels are always initialised to zero with the exception of the charge-dipole experiment, where we use the atomic dipole vector with additional parity and feature axes of size 1 (i.e., $\vec{\mu}\in\mathbb{R}^{1\times 3\times 1}$) to initialise the features with $\ell=1$ and odd parity as $\bm{x}^{(\ell_{-})} =\vec{\mu} W$ ($W \in \mathbb{R}^{1 \times H}$ is a learnable weight matrix that maps the dipole vector to the feature space). 

The $\ell=0$ components with even parity (invariant parts) of the final atomic features $\mathcal{X}^{[T]}$ are used to predict per-atom energies which are summed to give the total energy
\begin{align}
    E_{\text{ML}} = \sum_{m=1}^N w^\intercal \boldsymbol{x}_m^{[T](0_{+})} + E_{m,\text{bias}}\,,
\end{align}
where $w \in \mathbb{R}^H$ is a trainable vector, $E_{m, \text{bias}} \in \mathbb{R}$ is a trainable atom type dependent energy bias and $\boldsymbol{x}^{[T](0_{+})} \in \mathbb{R}^H$ is the invariant part of the final features at layer $T$ (parity and degree axes of size $1$ are omitted here for clarity). In the presence of atomic dipoles, the invariant part is obtained via a tensor product which maps the atomic features of all degrees to invariant features. Forces are calculated as the negative gradient of the energy \wrt the atomic positions ($\vec{F}_m = - \nabla_{\vec{r}_m} E$), which can be done efficiently using automatic differentiation.

\paragraph{Equivariant Message Passing Block}
Throughout this work, we use a prototypical equivariant message passing neural network (MPNN). It is build around equivariant continuous convolutions which are employed in many state-of-the-art MPNNs like NequIP,\cite{batzner20223} TFN,\cite{thomas2018tensor} and Equiformer.\cite{liao2022equiformer} The message sent from atom $n$ to atom $m$ is given by 
\begin{align}
    \boldsymbol{m}_{mn}^{[t]} = \left(\boldsymbol{W}(r_{mn}) \circ \boldsymbol{Y}(\hat{r}_{mn})\right) \Motimes_{L_{\text{MP}}}^{L_Y \, L_x} \boldsymbol{x}_n, \label{eq:so3-convolutions}
\end{align}
where $\boldsymbol{W}(r_{mn}) \in \mathbb{R}^{1\times(L_Y + 1)^2 \times H}$ is a learned radial filter function containing $H$ features for each degree $\ell$ up to $L_Y$. Importantly, the ``subslices'' of size $2\ell + 1$ for each degree $\ell$ are constrained to contain identical entries (otherwise the operation would not preserve equivariance). The spherical harmonics $\boldsymbol{Y}(\hat{r}_{mn}) \in \mathbb{R}^{1\times(L_Y + 1)^2\times1}$ use ``dummy axes'' of size $1$ (similar to the use of spherical harmonics in \autoref{eq:efa-update-methods}) and the element-wise multiplication `$\circ$' is broadcasted along these axes.  The messages from all neighbours $n$ in the neighbourhood $\mathcal{N}_{m}$ of atom $m$ (containing all atoms that satisfy $r_{mn} < r_{\text{cut}}$) are aggregated as $\boldsymbol{m}_m^{[t]} = \sum_{n \in \mathcal{N}_{m}} \boldsymbol{m}_{mn}^{[t]}$. Note that the radial filter $\boldsymbol{W}(r_{mn})$ in \autoref{eq:so3-convolutions} is enforced to smoothly decay to zero at $r_{\text{cut}}$, so that the learned representations vary continuously when atoms enter or leave $\mathcal{N}_{m}$. Finally, the updated atom representations are obtained as
\begin{align}
    \boldsymbol{x}_m^{[t+1]} = \text{MLP}\big[\boldsymbol{x}_m^{[t]} + \boldsymbol{m}_m^{[t]}\big],
\end{align}
where MLP is a multi-layer perceptron with two equivariant dense layers and gated SiLU\cite{elfwing2018sigmoid} non-linearity.

For standard MPNNs, each MP update consists of an equivariant MP Block. For an MPNN augmented with EFA, the atomic representations are passed to the equivariant MP block \textit{and} to the EFA block (see blow) and their outputs are added afterwards (see also \autoref{si:fig:mp-vs-mp-plus-efa}).

\paragraph{Euclidean Fast Attention Block}
Non-local atomic representations are calculated using the EFA update (\autoref{eq:efa-update-methods}), such that
\begin{align}
    \boldsymbol{m}_{m, \text{nl}}^{[t]} = \text{EFA}(\mathcal{X}^{[t-1]}, \mathcal{R})_m.
\end{align}
As for the MP block, per-atomic embeddings are refined via an equivariant MLP, such that
\begin{align}
    \boldsymbol{x}_{m, \text{nl}}^{[t+1]} = \text{MLP}\big[\boldsymbol{x}_m^{[t]} + \boldsymbol{m}_{m, \text{nl}}^{[t]}\big].
\end{align}

\subsection{Training}
Models are trained by minimising a combined loss of energy and forces
\begin{align}
    \mathcal{L} &= \frac{\lambda_E}{B}\sum_{b=1}^B \big(E_{b, \text{true}} - E_{b, \text{ML}}\big)^2 \nonumber\\
    &+ \frac{\lambda_F}{B}\sum_{b=1}^B \frac{1}{N_b}\sum_{m=1}^{N_b} \big|\big|\vec{F}_{m, \text{true}} - \vec{F}_{m, \text{ML}}\big|\big|_2^2,
\end{align}
where $B$ is the number of molecules per batch, $N_b$ is the number of atoms for molecule $b$ and $\lambda_E$ and $\lambda_F$ are scaling parameters for the energy and force components of the loss.

We use Adam\cite{kingma2014adam} for parameter optimisation (as implemented in the optax\cite{optax2020github} library) and an initial learning rate of $10^{-3}$. The learning rate is decayed to $10^{-5}$ at the end of training via an exponential decay schedule. In \autoref{si:tab:datasets-and-training} we report full information on optimisation settings for each experiment. 

\subsection{Model Hyperparameters}
If not stated otherwise, default EFA blocks employ a query and key dimension of $D_{qk} = 16$ and a value dimension of $D_v = 32$. The number of Lebedev grid points is set to $\numgridpoints = 50$. The maximal atomic separation varies between data sets and is reported in \autoref{si:tab:datasets-and-training}. For simplicity, we assume the degrees for queries, keys and values to always be equal and define the shorthand $L_\text{EFA} = L_{qk} = L_v$. Other model hyperparameters are described below.
\paragraph{Geometric Expressiveness} 
For the $k$-chains experiment, we use the equivariant MP block and the EFA block as standalone component. For the MP block, we follow the original publication\cite{joshi2023expressive} and use a hard cutoff of $r_\text{cut} = 10\,\an $ for simplicity. For the EFA block we use the hyperparameters outlined above. The final output consists of two values that are passed through softmax to predict a probability for each class. The models are trained by minimising the cross-entropy to the true class labels. The number of features for both blocks is $H = 32$ and the maximal degree in the MP block is $L_\text{max} = 2$. The number of MP updates is increased until the model is able to correctly classify the graphs. Because the maximal separation $r_\text{max}$ between atoms increases with $k$, the maximal frequency for ERoPE (see \autoref{eq:max-frequency-bound}) must be decreased accordingly,  leading to slower training convergence for the longest ($k=8$) chain. Based on theoretical analysis of positional encoding techniques\cite{tancik2020fourier} we can speed up convergence by increasing $b_\text{max} = 2\pi$ and adjust the number of Lebedev grid points to $\numgridpoints = 86$ accordingly (see \autoref{si:tab:bmax-to-lebedev-num}).
For the experiments probing distinguishability of neighbourhoods, we use an EFA update as described in \autoref{eq:euclidean-fast-attention-equivariant} and produce per degree invariants by taking the L2-norm per degree in the equivariant representation $\bm{x}$ of the central node. We compare the values for each degree $\ell$ for each pair of distinct neighbourhoods and label the smallest degree for which the invariants are different.

\paragraph{Two-Particle Systems} For the two-atom systems with a pairwise potential we use a feature dimension of $H = 128$; a maximal degree of $L_{\text{MP}} = 1$ is used in the MP block and $L_{\text{EFA}} = 0$ \rebuttal{and $\boldsymbol{\psi} = \mathrm{GELU}$ }in the EFA block. The spherical harmonics vector has maximal degree $L_Y = 0$ for all models expect for the one applied to the charge-dipole system, for which we additionally train a model with $L_Y = 1$ to show that the task can only be solved by including directional information (see main text).

\rebuttal{\paragraph{N-Particle System}}
\rebuttal{\label{rebuttal:methods:n-particle-systems} For the $N$-atom NaCl-like systems we use a feature dimension of $H = 32$, maximal degrees of $L_\text{MP} = 0$, and between $T=1$ and $T=5$ MP layers (see main text). For models with EFA, we use $L_\text{EFA}=0$, $\numgridpoints = 194$, $\omega_\text{max} = \frac{2\pi}{15}$, $D_{qk} = 128$, $D_v=8$ and $\boldsymbol{\psi} = \text{Id}$ as hyperparameters. Despite the systems having different maximal separation (sphere diameter), we employ the same value of $r_\text{max} = 30\,\an$ for all systems to maintain identical maximal frequency (see \autoref{eq:max-frequency-bound}).}

\rebuttal{
\paragraph{Non-local charge transfer}
For the non-local charge transfer benchmark, we distinguish between systems with and without PBCs. For the latter, we use $T = 3$ MP layers and a local cutoff of $r_\text{cut} = 4\,\an$, a feature dimension of $H = 128$ and $L_\text{MP} = 2$. For the model with EFA we use $L_\text{EFA} = 1$ \rebuttal{and $\boldsymbol{\psi} = \mathrm{GELU}$ }(other hyperparameters are set to the default values described above). We do not use the EFA block in the final layer as we found this to improve training stability and final accuracy.}

\rebuttal{For the periodic system we use $T = 2$ MP layers (all other hyperparameters for the MP model are the same as for the non-periodic systems). For EFA, we follow the protocol for PBCs described in SI \nameref{si:sec:other-symmetrisation-operations} and choose a maximal frequency of $\omega_\text{max} = \pi / 4$ in ERoPE with $L_\text{EFA} = 0$\rebuttal{ and $\boldsymbol{\psi} = \mathrm{GELU}$}.}

\rebuttal{
\paragraph{Materials}
For the models trained on the materials data from Ref.~\citenum{sauceda2022bigdml}, we use the same model as for the periodic system in the non-local charge transfer benchmark (see above).}

\paragraph{Molecular Systems}
For the molecular systems we use $T = 2$ for the S$_\text{N}$2 experiments and $T = 3$ for the dimer and cumulene results. The local cutoff is varied between $r_\text{cut} = 5\,\an$ and $r_\text{cut} = 10\,\an$ for the S$_\text{N}$2 experiments as described in the main text. We choose $r_\text{cut} = 4\,\an$ for the dimers and $r_\text{cut} = 3\,\an$ for cumulene. The feature dimension for the models with EFA block is $H=128$ for $T=2$ and $H=64$ for $T=3$. For models without EFA block, the feature dimension is increased to $H=162$ for $T=2$ and $H = 84$ for $T=3$, such that models with and without EFA block have approximately ($\leq 1\%$ deviation) the same number of parameters. The maximal degree in the MP block is $L_\text{MP} = 2$ and $L_\text{EFA} = 0$ in the EFA block, except for the cumulene structures where we additionally train models with $L_\text{EFA} = 1$ and $L_\text{EFA} = 2$ (see main text). The maximal degree of the spherical harmonics vector in the EFA update is always $L_Y = 0$. \rebuttal{All EFA models employ $\boldsymbol{\psi} = \mathrm{GELU}$.}

\subsection{Pairwise Potentials}
For the isotropic two-particle systems, the potential is given by
\begin{equation}
    V(r_{mn}) = \frac{1}{r_{mn}^3} - \frac{1}{r_{mn}},
\end{equation}
resembling attractive charge-charge interactions in the long-range regime. For the anisotropic two-particle system, we modify the long-range component to mimic charge-dipole interactions: One atom is assigned a ``charge'' and the other a unit ``dipole'' vector $\vec{\mu} \in \mathrm{R}^3$. The long-range potential is then given as 
\begin{equation}
    V(r_{mn}) = \frac{1}{r_{mn}^3} - \frac{\cos(\Theta)}{r_{mn}^2},
\end{equation}
where $\Theta$ is the angle between $\vec{\mu}$ and the normalised displacement vector between the atoms $\vec{r}_{mn} / r_{mn}$. The quadratic decay in distance $r$ reflects the physical behaviour of charge-dipole interactions.

\rebuttal{For the $N$-particle NaCl-like system, we assume Na to have positive $q = +1$ and Cl to have negative charge $q = -1$. The long-range interactions are described via a screened Coulomb potential
\begin{equation}
    V(r_{mn}) = k_e q_m q_n \frac{\mathrm{erf}(\alpha r_{mn})}{r_{mn}},
    \label{eq:damped-coulomb-potential}
\end{equation}
where $k_e$ is the Coulomb constant, $\mathrm{erf}$ is the error function and $\alpha = 0.5\,\an^{-1}$ is a scalar screening coefficient. The screening prevents diverging energy and forces when $r$ approaches zero and corresponds to the interaction of Gaussian charge densities.}

\rebuttal{\subsection{NaCl Data Creation}
We randomly place $N$ atoms within a sphere of diameter $d$. We consider systems of increasing diameter $d$, ranging from $10\,\an$ to $30\,\an$. We keep the density constant at 0.075\,atoms/$\an^3$, such that the number of atoms $N$ increases with $d$. To ensure a physically meaningful distribution of pairwise distances, we require that the distance $r_{mn}$ is $\geq (R^\text{vdW}_m + R^\text{vdW}_n) / 2$ for any two atoms $m$ and $n$ (we use $R^\text{vdW}_\text{Na}=0.95\,\an$ and $R^\text{vdW}_\text{Cl}=1.91\,\an$ as van der Waals radii).
}

\subsection{Dimer Data Subset Creation}
The SPICE data set\cite{eastman2023spice} serves as starting point for the dimer data. SPICE contains energies and forces for the dimer geometries in the DES370K data set\cite{donchev2021quantum} re-calculated using density functional theory (DFT). The resulting data set has 4612 entries and covers a variety of biologically relevant dimers. Each dimer consists of two molecules (monomers) which interact with each other via non-covalent (long-range) interactions. However, the generated data set is unbalanced in the sense that certain monomers appear more often within dimers than others, biasing the model towards a correct description of dimers with these monomers. To alleviate this problem, we create a curated subset of the dimers data which only uses the most frequent 9 monomers. This results in a new data set of 76 dimers.  Almost all combinations of monomers present in the original data set are also present in the curated version (76 vs.~81 possible combinations).
\section*{Data Availability}
The data for charge-dipole, NaCl clusters, SN2, cumulene, and dimers can be found at \url{https://doi.org/10.5281/zenodo.14750285}. The $k$-chains data is obtained via code at \url{https://github.com/chaitjo/geometric-gnn-dojo}. The data for distinguishability of local neighbourhoods has been taken from \url{https://github.com/google-research/e3x/blob/main/tests/nn/modules_test.py} (starting from line 315) and was originally proposed in Ref.~\citenum{pozdnyakov2020incompleteness}. \rebuttal{The data for the non-local charge transfer benchmark is taken from Ref.~\citenum{ko2021fourth} and a preprocessed version of it for use with EFA can be found in the Zenodo repository. The data for \autoref{tab:bigdml} is taken from Ref.~\citenum{sauceda2022bigdml}.}

\section*{Code Availability}
An implementation of the Euclidean fast attention block is publicly available at \url{https://github.com/thorben-frank/euclidean_fast_attention}. The repository further includes code for data loading and processing, a reference implementation of an O(3)-equivariant message passing neural network, as well as code for training and evaluation.
\section*{Acknowledgements}
This work was in part supported by the German Ministry for Education and Research (BMBF) under Grants 01IS14013A-E, 01GQ1115, 01GQ0850, 01IS18025A, 031L0207D, and 01IS18037A. K.R.M.\ was partly supported by the Institute of Information \& Communications Technology Planning \& Evaluation (IITP) grants funded by the Korea government (MSIT) (No. 2019-0-00079, Artificial Intelligence Graduate School Program, Korea University and No. 2022-0-00984, Development of Artificial Intelligence Technology for Personalized Plug-and-Play Explanation and Verification of Explanation). We thank Stefan Blücher and Hartmut Maennel for helpful comments on the manuscript.
\bibliography{main}

\appendix
\newpage
\onecolumn
\begin{center}
\textbf{\large Supplementary Information}
\end{center}
\begin{center}
\textbf{\large Euclidean Fast Attention - Machine Learning Global Atomic \\ Representations at Linear Cost}
\end{center}

\renewcommand{\thefigure}{S\arabic{figure}}
\setcounter{figure}{0}
\renewcommand{\thetable}{S\arabic{table}}
\setcounter{table}{0}
\renewcommand{\theequation}{S\arabic{equation}}
\setcounter{equation}{0}

\section{Attention and Euclidean data}
\label{sec:attention-and-euclidean-data}
In the following, we briefly describe standard (quadratically-scaling) self-attention\cite{vaswani2017attention} and linear-scaling variants. Further, we outline how standard attention can be extended for data embedded in Euclidean space, and why a similar modification is difficult for linear-scaling attention formulations.

Given a set of $N$ features $\mathcal{X} = \{\boldsymbol{x}_1,\dots,\boldsymbol{x}_N\,|\,\boldsymbol{x}_m\in\mathbb{R}^H\}$, self-attention calculates
\begin{equation}
    \mathrm{ATT}(\mathcal{X})_m = \frac{\sum_{n=1}^N \mathrm{sim}(\boldsymbol{q}_m, \boldsymbol{k}_n) \, \boldsymbol{v}_n}{\sum_{n=1}^N\mathrm{sim}(\boldsymbol{q}_m, \boldsymbol{k}_n)}
    =\sum_{n=1}^N a_{mn} \boldsymbol{v}_n \,, 
    \label{eq:quadratic-scaling-attention}
\end{equation}
where
\begin{equation*}
\begin{aligned}
a_{mn} = \frac{\mathrm{sim}(\boldsymbol{q}_m, \boldsymbol{k}_n)} {\sum_{n=1}^N\mathrm{sim}(\boldsymbol{q}_m, \boldsymbol{k}_n)}
\end{aligned}
\end{equation*}
are so-called attention coefficients. The vectors $\boldsymbol{q},\boldsymbol{k} \in \mathbb{R}^{D_{qk}}$, and $\boldsymbol{v}\in \mathbb{R}^{D_{v}}$ are called query, key, and value, respectively, and are obtained from the features $\boldsymbol{x}$, typically via linear transformations
\begin{equation*}
\boldsymbol{q} = \boldsymbol{W}_q \boldsymbol{x}\qquad
\boldsymbol{k} = \boldsymbol{W}_k \boldsymbol{x}\qquad
\boldsymbol{v} = \boldsymbol{W}_v \boldsymbol{x}
\end{equation*}
with trainable weight matrices $\boldsymbol{W}_q,\boldsymbol{W}_k \in \mathbb{R}^{D_{qk} \times H}$ and  $\boldsymbol{W}_v \in \mathbb{R}^{D_{v} \times H}$. The similarity kernel is usually chosen as 
\begin{equation}
    \mathrm{sim}(\boldsymbol{q}, \boldsymbol{k}) = \exp\left(\frac{\boldsymbol{q}^\T \boldsymbol{k}}{\sqrt{D_{qk}}}\right)\,.
    \label{eq:exponential-kernel}
\end{equation}

To apply self-attention to structures embedded in Euclidean space, \autoref{eq:quadratic-scaling-attention} needs to be modified to also include information about the positions $\mathcal{R} = {\{\vec{r}_1, \dots, \vec{r}_N\,|\,\vec{r}_m \in \mathbb{R}^3\}}$ associated with the features $\mathcal{X}$. A straightforward way to achieve this is to define a geometric version of self-attention
\begin{align}
    \text{ATT}_{\text{Geom}}(\mathcal{X}, \mathcal{R})_m = \frac{\sum_{n=1}^N \text{sim}(\boldsymbol{q}_m, \boldsymbol{k}_n, r_{mn}) \, \boldsymbol{v}_n}{\sum_{n=1}^N\text{sim}(\boldsymbol{q}_m, \boldsymbol{k}_n, r_{mn})}\,, \label{eq:quadratic-scaling-geometric-attention}
\end{align}
which uses a modified similarity kernel that explicitly depends on the pairwise distances $r_{mn} = \lVert\vec{r}_m - \vec{r}_n\rVert$. The use of distances to encode spatial information is convenient, because the resulting operation is naturally invariant to rigid translations and rotations, and therefore respects physical symmetries. A possible choice would be
\begin{equation}
    \mathrm{sim}(\boldsymbol{q}, \boldsymbol{k},  r) = \exp\left(\frac{\boldsymbol{q}^\T \boldsymbol{k} \cdot f(r)}{\sqrt{D_{qk}}}\right)\,,
    \label{eq:geometric-exponential-kernel}
\end{equation}
which is similar to \autoref{eq:exponential-kernel}, except that the dot product $\boldsymbol{q}^\T \boldsymbol{k}$ is modulated by a (possibly learned) function $f$ of the distance $r$.

Note that \autoref{eq:quadratic-scaling-geometric-attention} can also be thought of as a special type of message passing (MP) layer acting on a fully-connected graph, where the ``message'' sent from node $n$ to node $m$ is given by $a_{mn}\boldsymbol{v}_n$. Given this analogy, it is apparent that the overall complexity of evaluating \autoref{eq:quadratic-scaling-geometric-attention} is $\mathcal{O}(N^2)$. It would be possible to introduce a cutoff distance $r_{\text{cut}}$ and recover $\mathcal{O}(N)$ scaling, but then \autoref{eq:quadratic-scaling-geometric-attention} would not be able to model global correlations anymore and have the same issues as ordinary message passing with a local cutoff (see \nameref{subsec:limitations-of-local-models}).

However, it is also possible to derive a linear-scaling version of self-attention\cite{katharopoulos2020transformers} in an alternative manner by re-writing the similarity kernel (\autoref{eq:exponential-kernel}) as a scalar product in an associated (implicit) feature space\cite{mercer1909functions,aronszajn1950theory,scholkopf1998nonlinear,scholkopf1999input,mueller2001,scholkopf2002learning} as
\begin{equation}
    \mathrm{sim}(\boldsymbol{q}, \boldsymbol{k}) = \boldsymbol{\psi}(\boldsymbol{q})^\T \boldsymbol{\psi}(\boldsymbol{k})
    \label{eq:similarity_kernel_rewrite}
\end{equation} via a feature map $\boldsymbol{\psi}$ taking values in a separable Hilbert space of dimensions $D_\psi \leq \infty$. Inserting \autoref{eq:similarity_kernel_rewrite} into \autoref{eq:quadratic-scaling-attention} leads to
\begin{subequations}
\begin{align}
    \mathrm{ATT}(\mathcal{X})_m &= \frac{\sum_{n=1}^N \boldsymbol{\psi}(\boldsymbol{q}_m)^\T \boldsymbol{\psi}(\boldsymbol{k}_n) \, \boldsymbol{v}_n}{\sum_{n=1}^N \boldsymbol{\psi}(\boldsymbol{q}_m)^\T \boldsymbol{\psi}(\boldsymbol{k}_n)}\,,
    \label{eq:linear-scaling-attention-still-quadratic}
    \\
    \intertext{which can equivalently be written as}
    \mathrm{ATT}_{\text{Lin}}(\mathcal{X})_m &=  \frac{\boldsymbol{\psi}(\boldsymbol{q}_m)^\T\sum_{n=1}^N  \boldsymbol{\psi}(\boldsymbol{k}_n) \, \boldsymbol{v}_n^\T}{\boldsymbol{\psi}(\boldsymbol{q}_m)^\T \sum_{n=1}^N \boldsymbol{\psi}(\boldsymbol{k}_n)}\,.
    \label{eq:linear-scaling-attention}
\end{align}
\end{subequations}
In this formulation, it is evident that the sums $\sum_{n=1}^N \boldsymbol{\psi}(\boldsymbol{k}_n) \, \boldsymbol{v}_n^\T$ and $\sum_{n=1}^N \boldsymbol{\psi}(\boldsymbol{k}_n)$ are identical for each query vector $\boldsymbol{q}_m$ and therefore only need to be computed once. Thus, the overall complexity of computing self-attention for every input via \autoref{eq:linear-scaling-attention} is only $\mathcal{O}(N)$ (see \autoref{fig:overview:quadratic-vs-linear-attention} for an illustration that contrasts the evaluation of \autoref{eq:linear-scaling-attention-still-quadratic} with \autoref{eq:linear-scaling-attention}).

\rebuttal{\label{rebuttal:formal_exactness}We note that while Eq.~\ref{eq:similarity_kernel_rewrite} is formally exact and re-writing Eq.~\ref{eq:quadratic-scaling-attention} in the form of Eq.~\ref{eq:linear-scaling-attention} is always possible in theory, a practical implementation is only feasible when $D_{\psi}$ is finite, and only useful when $D_{\psi} < N$. Since the associated feature space of the exponential kernel (Eq.~\ref{eq:exponential-kernel}) is infinite-dimensional, it either needs to be approximated,\cite{choromanski2020rethinking} or an entirely different kernel function (feature map) needs to be chosen. Fortunately, for many kernels (e.g., the polynomial kernel), the associated feature space already is finite-dimensional and does not need to be approximated. Instead of trying to find an (approximate) feature map for a specific kernel, it is also possible to ``invert the process'' and simply choose an (arbitrary) finite-dimensional feature map, which is what we do for EFA.}

To apply linear-scaling attention to Euclidean data, we would need to re-write the geometric version of self-attention (\autoref{eq:quadratic-scaling-geometric-attention}) in the form of \autoref{eq:linear-scaling-attention}. This would require re-writing the modified similarity kernel as
\begin{equation}
\mathrm{sim}(\boldsymbol{q}_m, \boldsymbol{k}_n, r_{mn}) = \boldsymbol{\psi}(\boldsymbol{q}_m, \vec{r}_m)^\T \boldsymbol{\psi}(\boldsymbol{k}_n, \vec{r}_n)\,,
\label{eq:hypothetical_geometric_similarity_kernel_rewrite}
\end{equation}
but it is non-obvious how to design a feature map $\boldsymbol{\psi}$ that achieves this. In fact, $\mathcal{O}(N)$ scaling seems inherently incompatible with encoding information about all pairwise distances $r_{mn}$, because computing them directly has quadratic complexity. A possible alternative could be to encode geometric information in some other way that does not rely on pairwise distances, but then it is unclear how to ensure that physical symmetries, such as translational and rotational invariance, are respected by the resulting mechanism. Our \nameref{sec:euclidean-fast-attention} method addresses these challenges (see main text).

\phantomsection
\section{Analytic Solution of the Surface Integral} \label{si:sec:surface-integral}
In the following we give a step-by-step solution of the surface integral (see \autoref{eq:surface-integral-solution})
\begin{equation*}
I = \frac{1}{4\pi}\int_{S^2} e^{i \omega \vec{u}\cdot\vec{r}_{mn}} \ \mathrm{d}\vec{u}\,.
\end{equation*}
We start by re-writing the integral in spherical coordinates
\begin{equation*}
I = \frac{1}{4\pi}\int_{0}^{2\pi} \int_{0}^{\pi} e^{i \omega r_{mn} \cos({\theta})} \sin(\theta)\ \mathrm{d}\theta\ \mathrm{d}\phi\,,
\end{equation*}
where we have substituted $\vec{u} = [\sin(\theta)\cos(\phi)\ \sin(\theta)\sin(\phi)\ \cos(\theta)]^\T$
and assumed (without loss of generality) that $\vec{r}_{mn}$ is aligned with the $z$-axis of our coordinate system, such that $\vec{u}\cdot\vec{r}_{mn} = r_{mn} \cos({\theta})$ (with $r_{mn}=\lVert\vec{r}_{mn}\rVert$). Next, we substitute $x \coloneqq -\cos(\theta)$ such that $\sin(\theta)\ \mathrm{d}\theta=\mathrm{d}x$ and $\int_{0}^{\pi} \rightarrow \int_{-1}^{1}$, giving
\begin{equation*}
I = \frac{1}{4\pi}\int_{0}^{2\pi} \int_{-1}^{1} e^{-i \omega r_{mn} x}\ \mathrm{d}x\ \mathrm{d}\phi\,.
\end{equation*}
Using the analytic solution for
\begin{equation*}
\int_{-1}^{1} e^{-i a x}\ \mathrm{d}x = 2\frac{\sin(a)}{a} = 2\, \mathrm{sinc}(a)\,,
\end{equation*}
we arrive at the solution
\begin{equation*}
I = \frac{1}{2\pi} \int_{0}^{2\pi} \mathrm{sinc}(\omega r_{mn}) \ \mathrm{d}\phi = \mathrm{sinc}(\omega r_{mn})\,.
\end{equation*}


For the general case with $\ell\geq0$, recall the encoding map used in ERoPE (see \nameref{sec:methods}), which is given as
\begin{align}
    \varphi_{km\vec{u}} \coloneqq c_{km} e^{i \omega_k \vec{u} \cdot \vec{r}_m}.
\end{align}
The dot product for encoding maps at atomic positions $\vec{r}_m$ and $\vec{r}_n$ is given as
\begin{align}
    \braket{\varphi_{km\vec{u}}, \varphi_{kn\vec{u}}} = c_{kmn} e^{i \omega_k \vec{u}\cdot\vec{r}_{mn}},
    \label{si:eq:encoding_map_dot}
\end{align}
where $c_{kmn} \coloneqq c_{km}\compconj{c}_{kn} \in \mathbb{C}$. The complex exponential can be written in terms of spherical harmonic functions using plane wave expansion
\begin{align}
    e^{i \omega_k \vec{u} \cdot \vec{r}_{mn}} = 4\pi  \sum_{\ell=0}^{\infty} \sum_{M=-\ell}^\ell i^\ell j_{\ell}(\omega_k r_{mn}) \compconj{Y}_{\ell}^M(\vec{u}) Y_{\ell}^M(\hat{r}_{mn}),
    \label{si:eq:plane-wave-expansion}
\end{align}
where $j_\ell$ is the $\ell$th Bessel function and $Y_l^M$ is a spherical harmonic of degree $\ell$ and order $M$. The integral for a single component of order $M$ and degree $\ell$ in the spherical harmonics vector $\boldsymbol{Y}$ in \autoref{eq:euclidean-fast-attention-equivariant} is given as
\begin{align}
    I = \frac{1}{4\pi}\int_{S^2} \mathrm{d}\vec{u} \braket{\varphi_{km\vec{u}}, \varphi_{kn\vec{u}}} Y_{\tilde{\ell}}^{\widetilde{M}}(\vec{u}).
\end{align}
Using \autoref{si:eq:plane-wave-expansion}, a compact expression for the integral 
\begin{align}
    I_{\tilde{\ell}}^{\widetilde{M}} &= c_{kmn} \, \sum_{\ell=0}^{\infty} \sum_{M=-\ell}^\ell i^\ell \, \int_{S^2} \mathrm{d}\vec{u} j_{\ell}(\omega_k r_{mn}) \compconj{Y}_{\ell}^M(\vec{u}) Y_{\ell}^M(\hat{r}_{mn}) Y_{\tilde{\ell}}^{\widetilde{M}}(\vec{u})\\
    &= c_{kmn} \, \sum_{\ell=0}^{\infty} \sum_{M=-\ell}^\ell i^\ell \, j_{\ell}(\omega_k r_{mn}) Y_{\ell}^M(\hat{r}_{mn}) \delta_{M\widetilde{M}, \ell \tilde{\ell}} \\
    &= i^{\tilde{\ell}}\, c_{kmn} j_{\tilde{\ell}}(\omega_k r_{mn}) Y_{\tilde{\ell}}^{\widetilde{M}}(\hat{r}_{mn})
\end{align}
can be obtained. Here we used the orthonormality relation of the spherical harmonics $\int_{S^2}\mathrm{d}\vec{u} \, \compconj{Y}_{\tilde{\ell}}^{\widetilde{M}} Y_{\ell}^{M} = \delta_{M\widetilde{M}, \ell\tilde{\ell}}$. Evaluation of the integral for all orders $M = -\ell, \dots, +\ell$ for a given degree $\ell$ in $\boldsymbol{Y}$ and subsequent concatenation can be written using the tensor product 
\begin{align}
    I_{\tilde{\ell}} &= \frac{1}{4\pi} \int_{S^2} \mathrm{d}\vec{u} \braket{\varphi_{km\vec{u}}, \varphi_{kn\vec{u}}} \otimes \boldsymbol{Y}_{\tilde{\ell}} \\
    &= i^{\tilde{\ell}}\, c_{kmn} j_{\tilde{\ell}}(\omega_k r_{mn}) \otimes \boldsymbol{Y}_{\tilde{\ell}}(\hat{r}_{mn}), \label{si:eq:surface-integral-solution-general-case}
\end{align}
where $\boldsymbol{Y}_\ell$ is a vector containing all spherical harmonics with degree $\ell$ in $\boldsymbol{Y}$. Repeating this for all different $\ell = 0, \dots \ell_\text{max}$ in $\boldsymbol{Y}$ puts equivariant SO(3) convolutions (\autoref{eq:so3-convolutions-main}) into relation to the equivariant EFA update, as stated in main text. It should be noted, that, for the convenience of notation and to highlight the fundamental relation between the dot product and the spherical harmonics, the above derivation assumed $\boldsymbol{x}$ to be an invariant representation ($\ell = 0$). In the setting of invariant features, the output of EFA equals equivariant filters (as, e.g., used in SpookyNet~\cite{unke2021spookynet}). However, EFA similarly works (and is actually implemented) for equivariant features $\boldsymbol{x}$, where technical implementation details are explained in the \nameref{sec:methods} section. This corresponds to full $\mathrm{SO}(3)$ convolutions as found in many equivariant MPNNs (as in, e.g., NequIP\cite{batzner20223}).

For the invariant case $\ell = 0$, the spherical harmonics vector $\boldsymbol{Y}_\ell$ is simply a scalar $1$ (when using Racah normalisation) and \autoref{si:eq:surface-integral-solution-general-case} simplifies to 
\begin{align}
    I_{0} = c_{kmn} j_0(\omega_k r_{mn})\,.
\end{align}
Note that $j_0$ is identical to the sinc function, so the general solution correctly recovers the integral solution for the invariant case from above.

\begin{figure*}
    \centering
     \includegraphics[width=\textwidth]{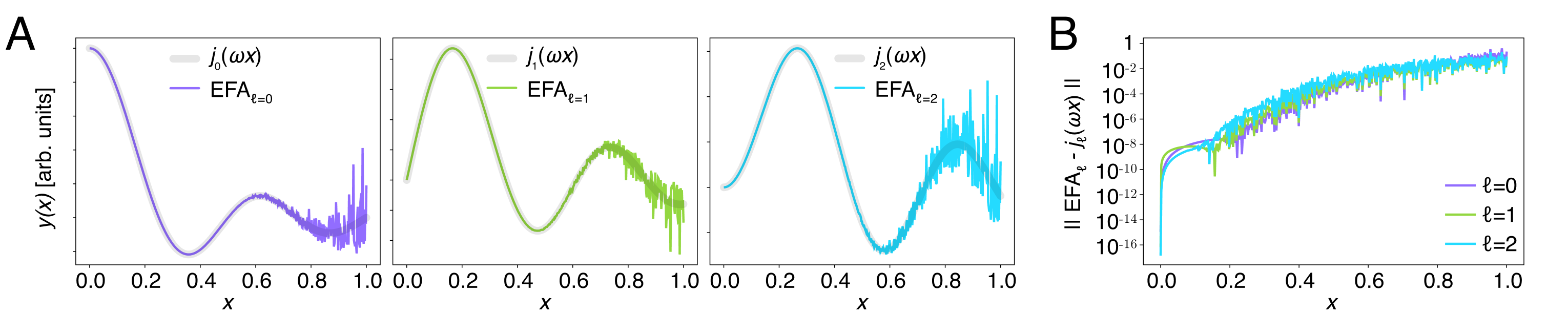}
    \begin{subfigure}{\linewidth}
        \phantomcaption{}
        \label{si:fig:efa-vs-bessel:function}
    \end{subfigure}
    \begin{subfigure}{\linewidth}
        \phantomcaption{}
        \label{si:fig:efa-vs-bessel:deviation}
    \end{subfigure}
    \caption{\textbf{Comparison of Euclidean fast attention (EFA) with the analytic integral solution. }\textbf{(A)} Comparison of the output of EFA and the first three Bessel functions $j_\ell$ for increasing degree $\ell$, which are the analytical solution of the surface integral (\autoref{si:eq:surface-integral-solution-general-case}). \textbf{(B)} Deviation between the Bessel function and the output of EFA as function of $x$. For the frequency we chose $\omega = 4\pi$ and the number of Lebedev grid points was set to $\numgridpoints = 50$.} 
    \label{si:fig:efa-vs-bessel}
\end{figure*}

We verify numerically, that the output of Euclidean fast attention recovers the functional form of the first three Bessel functions (see \autoref{si:fig:efa-vs-bessel:function}). As long as the integration grid is sufficiently large, the output is equivalent to the expected functional behaviour up to numerical precision (see \autoref{si:fig:efa-vs-bessel:deviation}). The larger the degree $\ell$ in $\boldsymbol{Y}_\ell$, the sooner the output starts to deviate from the exact solution. However, the effect of increasing $\ell$ only marginally reduces the threshold distance when numerical precision (here $10^{-5}$) is violated. The driving source for deviation from the exact solution remains the maximum distance in the data. 

\phantomsection
\section{Other Symmetrisation Operations} \label{si:sec:other-symmetrisation-operations}
As argued in the main text, evaluating the product of encoding maps (see \autoref{si:eq:encoding_map_dot})
\begin{align}
    \braket{\varphi_{km\vec{u}}, \varphi_{kn\vec{u}}} \sim e^{i \omega_k \vec{u} \cdot \vec{r}_{mn}} = e^{i \omega_k r_{mn} \cos(\theta)},
\end{align}
depends on the angle $\theta$ between vector $\vec{u}$ and displacement vector $\vec{r}_{mn} = \vec{r}_m - \vec{r}_n$, which is sensitive to the choice of $\vec{u}$, effectively defining a reference frame. Thus, the output depends on the orientation $\theta$ between positions and $\vec{u}$, which results in representations that are neither invariant nor equivariant under rigid rotation of all input positions. To obtain an invariant/equivariant operation, we typically integrate over the unit sphere (see \autoref{eq:surface-integral-solution}), but other symmetrisation operations are possible and may be preferable depending on the application at hand.

\paragraph{Lattice Vectors}
In the setting of periodic systems, lattice vectors can be used to define a reference frame as e.g.~done in Ewald message passing (MP).\cite{kosmala2023ewald} The symmetrisation of the EFA update under periodic boundary conditions can be written as (assuming a maximal degree $L_Y = 0$ for the spherical harmonics for simplicity)
\begin{align}
    \text{EFA}_\text{lattice}(\mathcal{X}) = \sum_{p=1}^3 \boldsymbol{B}_{\vec{u}_p},
    \label{eq:pbc_symmetrisation}
\end{align}
where $\vec{u}_p = \vec{l}_p$ and $\vec{l}_p \in \big\{ \vec{l}_1, \vec{l}_2, \vec{l}_3\,|\,\vec{l}_p \in \mathbb{R}^3\big\}$ are the lattice vectors and $\boldsymbol{B}_{\vec{u}_p}$ is defined as in \autoref{eq:efa-B-tensor}. This ensures invariance \wrt~rotations of the unit cell as a whole (rotational symmetry within the unit cell is broken, which is a natural assumption for periodic systems). \rebuttal{Other choices for $\vec{u}_p$ are also possible, i.e., normalized or reciprocal lattice vectors, which might be better suited when training on different unit cells simultaneously.}

\paragraph{Canonicalisation of the Input Geometry}
In settings without periodic boundary conditions, it is still possible to use \autoref{eq:pbc_symmetrisation} by defining the vectors $\vec{u}_p \coloneq \vec{b}_p$ as the right singular vectors $\{\vec{b}_1, \vec{b}_2, \vec{b}_3 \,|\, \vec{b}_p \in \mathbb{R}^3 \}$ of the singular value decomposition (SVD) of the input geometry $\boldsymbol{R} \in \mathbb{R}^{N\times 3}$. However, SVD is only defined up to a sign and basis vectors are non-unique for degenerate eigenvalues, which appear for highly symmetric systems. This option is thus less general than integration over the unit sphere and only useful in specialised settings.

\phantomsection
\section{\rebuttal{Extended Scaling Analysis}}\label{si:sec:extended-scaling-analysis}
\rebuttal{In this section, we empirically investigate the computational cost and accuracy of EFA when scaled to increasingly large system sizes to confirm they match theoretical expectations. For this, we consider a purely EFA-based energy predictor (without any other ML model) in order to be able to investigate the EFA mechanism in isolation. The total energy is modelled as 
\begin{equation}
    E_\text{EFA} = \int_{S^2}\left(- E_{\text{self}, \vec{u}} + \sum_{m=1}^N E_{m, \vec{u}} \right)\,\mathrm{d}\vec{u}\,, 
    \label{si:eq:efa-energy-predictor}
\end{equation}
where 
\begin{align}
    E_{m, \vec{u}} &= \mathrm{ERoPE}_{\vec{u}}\left( \vec{r}_m, q_m \cdot \boldsymbol{c} \right) \cdot \sum_{n = 1}^N \mathrm{ERoPE}_{\vec{u}}\left( \vec{r}_n, q_n \cdot \boldsymbol{c} \right)\,,\\
    E_{\text{self}, \vec{u}} &= \sum_{n=1}^N \lVert\mathrm{ERoPE}_{\vec{u}}(\vec{r}_n, \boldsymbol{c})\rVert_2^2\,,
\end{align}  
$\boldsymbol{c} \in \mathrm{R}^D$ is a trainable coefficient vector, and $q_m \in \mathbb{R}$ denotes the charge of the $m$-th atom. This is a special case of the more general EFA mechanism described in the main text with $\boldsymbol{q}_m = q_m \cdot \boldsymbol{c}$, $\boldsymbol{k}_n = q_n \cdot \boldsymbol{c}$, and $\boldsymbol{v}_n = 1$. The term $E_{\text{self}, \vec{u}}$ corresponds to a ``self-interaction energy'', which we explicitly subtract here so that only pairwise interactions with $m\neq n$ contribute to $E_\text{EFA}$. Note that this is not necessary when EFA is combined with MPNNs or other ML models, because the self-interaction energy corresponds to a constant energy shift that can be easily accounted for by learnable bias terms. By varying the coefficients $\boldsymbol{c}$, this model essentially learns a specific radial interaction kernel. For example, for modelling electrostatics, optimal coefficients $\boldsymbol{c}^*$ would correspond to $\int_{S^2} \mathrm{ERoPE}_{\vec{u}}(\vec{r}_m, \boldsymbol{c}^*) \mathrm{ERoPE}_{\vec{u}}(\vec{r}_n, \boldsymbol{c}^*)\, \mathrm{d}\vec{u} \propto r_{mn}^{-1}$. As such, \autoref{si:eq:efa-energy-predictor} can only fit simple pairwise potential energy functions of the form given in \autoref{eq:pairwise-potential}. However, \emph{if} a potential is well approximated in terms of pairwise interactions, suitable coefficients $\boldsymbol{c}$ should be learnable from small systems and generalise to a much larger number of atoms without accuracy degradation.}

\begin{figure*}[t!]
    \centering
    \includegraphics[width=\linewidth]{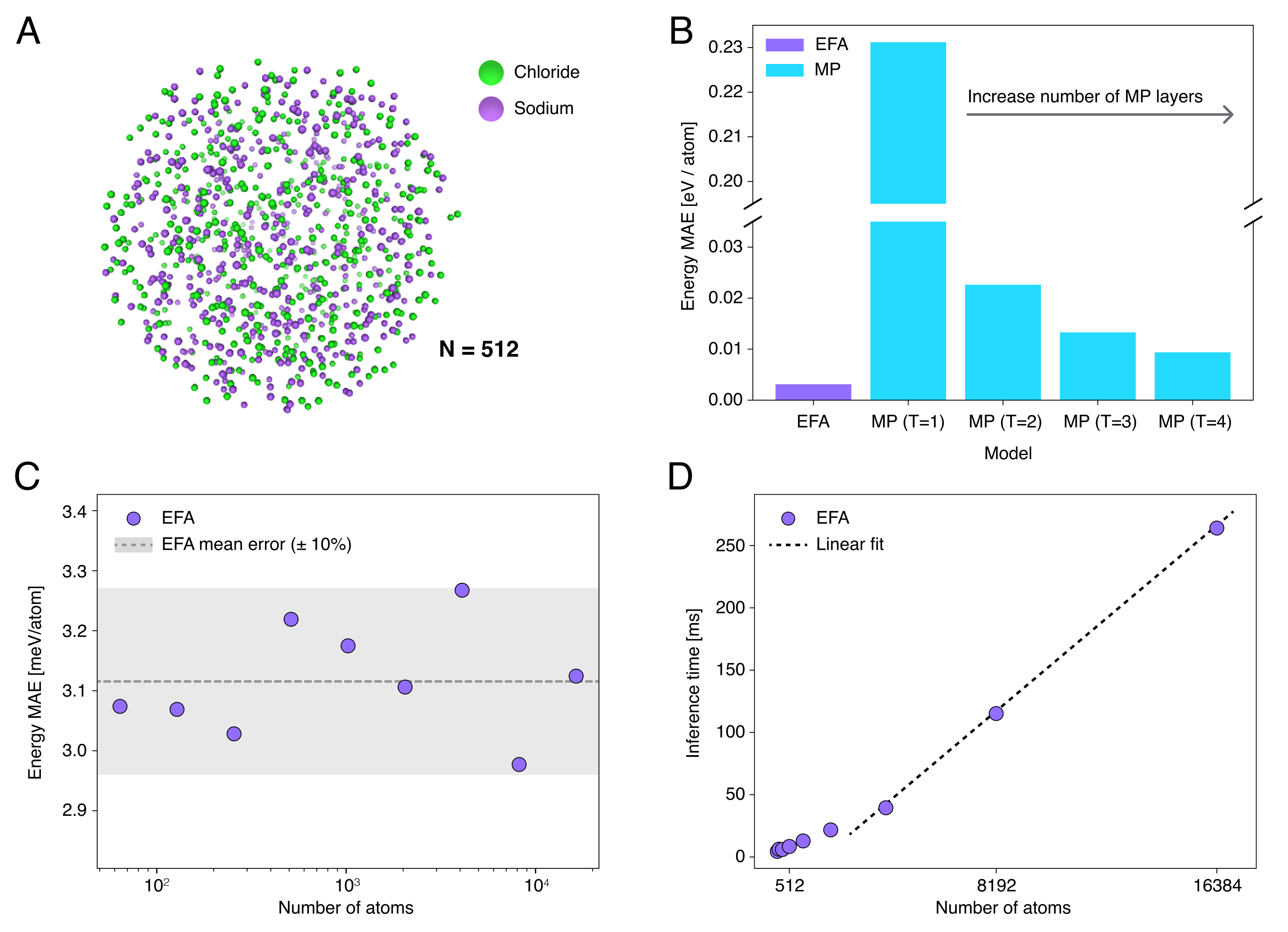}
    \begin{subfigure}{\linewidth}
        \phantomcaption{}
        \label{si:fig:results-panel-NaCl-toy:structure-image}
    \end{subfigure}
    \begin{subfigure}{\linewidth}
        \phantomcaption{}
        \label{si:fig:results-panel-NaCl-toy:mp-vs-efa}
    \end{subfigure}
    \begin{subfigure}{\linewidth}
        \phantomcaption{}
        \label{si:fig:results-panel-NaCl-toy:number-of-atoms-vs-error}
    \end{subfigure}
    \begin{subfigure}{\linewidth}
        \phantomcaption{}
        \label{si:fig:results-panel-NaCl-toy:number-of-atoms-vs-time}
    \end{subfigure}
    \caption{\rebuttal{\textbf{Scaling analysis for NaCl-like systems.} \textbf{(A)} Visual illustration of $N = 512$ particles distributed in a sphere of diameter $50\,\an$. \textbf{(B)} Mean absolute error (MAE) of energy predictions for Euclidean fast attention (EFA) compared to an MPNN with $r_\text{cut}=15\,\an$ (for an increasing number of layers $T$). Note that for $T=4$, the effective cutoff is $r_\text{eff}=60\,\an$ and covers the whole system. \textbf{(C)} Energy MAE of EFA for an increasing number of atoms. The dashed line is the mean over all investigated system sizes and the grey shaded area is the $\pm10\%$ interval around the mean. As expected, the prediction error is independent of system size. \textbf{(D)} Time per energy calculation for EFA as function of number of atoms, measured on Nvidia H100 GPU. We hypothesise that the observed sublinear scaling behaviour for small system sizes is due to the calculation not being compute-bound in this regime.}}
    \label{si:fig:results-panel-NaCl_toy}
\end{figure*}

\rebuttal{To empirically confirm this property of EFA, we consider the NaCl-like model potential from the main text (see \autoref{eq:damped-coulomb-potential}) and train only on two-atom systems with pairwise distances sampled uniformly between $0\,\an$ and $60.5\,\an$ to find the coefficient vector $\boldsymbol{c}$ (we set $q_\text{Na} = +1$ and $q_\text{Cl} = -1$). We then evaluate the trained model on systems with between $N=64$ up to $16,384$ atoms (distributed in a sphere of diameter $50\,\an$, see \autoref{si:fig:results-panel-NaCl-toy:structure-image}). As expected, we find that the energy prediction error per atom is independent of system size (\autoref{si:fig:results-panel-NaCl-toy:number-of-atoms-vs-error}) and inference time scales linearly in the number of atoms (\autoref{si:fig:results-panel-NaCl-toy:number-of-atoms-vs-time}). For completeness, we also confirm that an MPNN with $r_\text{cut}=15\,\an$ has significantly larger prediction errors, even when the number of MP layers $T$ is increased until the effective cutoff covers the whole system (\autoref{si:fig:results-panel-NaCl-toy:mp-vs-efa}). As hyperparameters for this experiment, we choose $D = 128$ as the size of the learnable coefficient vector $\boldsymbol{c}$, set the number of grid points for the numerical integration to $\numgridpoints = 2000$, and use a maximum frequency of $\omega_\text{max} = \pi$ for ERoPE.}

\begin{figure}
    \centering
    \includegraphics[width=0.99\linewidth]{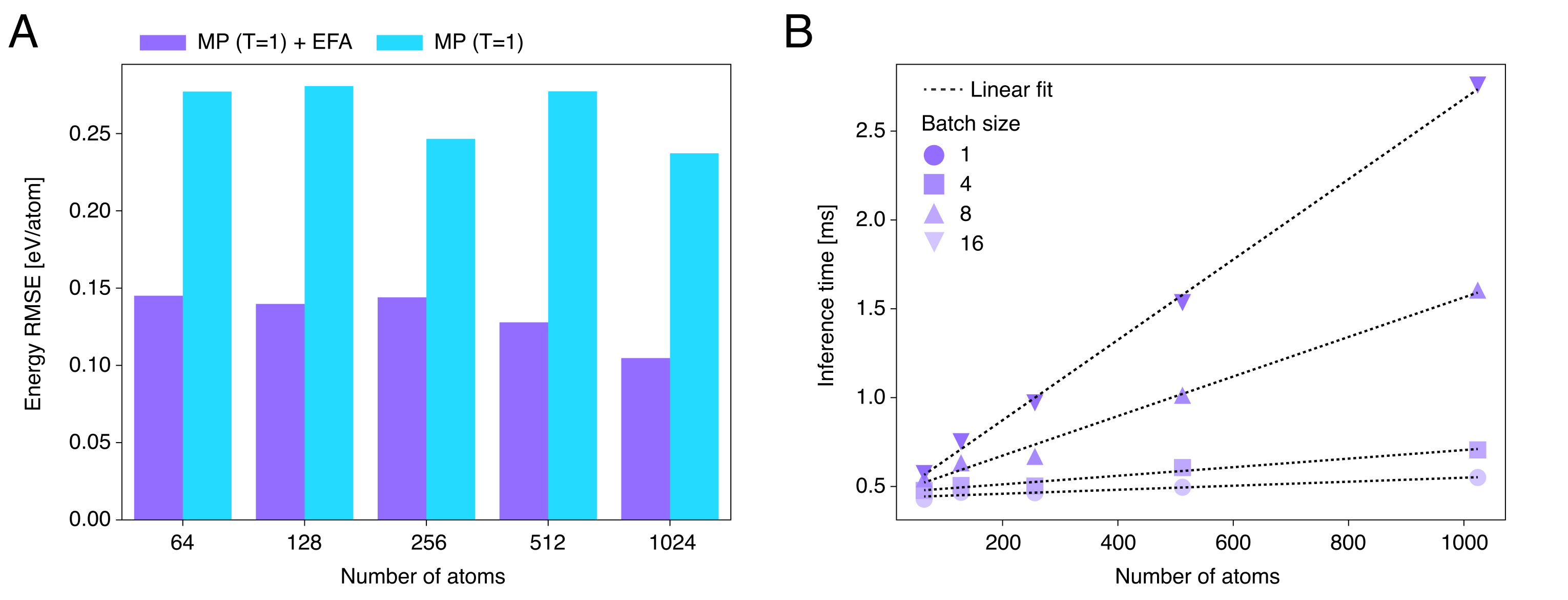}
    \caption{\rebuttal{\textbf{Scaling analysis for periodic NaCl-like systems.} \textbf{(A)} Energy root mean square error (RMSE) for an increasing number of atoms for periodic NaCl system. \textbf{(B)} Inference time as a function of number of atoms for different batch sizes. Runtimes were measured on Nvidia H100.}}
    \label{si:fig:results-panel-NaCl_toy-pbc}
\end{figure}

\rebuttal{
\paragraph{Scaling under periodic boundary conditions (PBCs)}
For completeness, we also investigate whether linear scaling \wrt the number of atoms holds for the periodic version of EFA (see \nameref{si:sec:other-symmetrisation-operations} for details). For this, we consider periodic NaCl-like toy systems, where instead of placing the atoms within a sphere of a given diameter, they are placed within a cubic unit cell of side length $L = 25\,\an$. The reference (ground truth) energy (and forces) are calculated via standard Ewald summation. We then train an MP model with $T = 1$ layer as well as an EFA-augmented version, using the same model hyperparameters as for the spherical NaCl systems in the main body of the text. Similar to the results for non-periodic systems in \autoref{si:fig:results-panel-NaCl_toy}, we find that accuracy does not decrease and evaluation costs scale linearly with system size (\autoref{si:fig:results-panel-NaCl_toy-pbc}).}

\section{Relation to Ewald Message Passing}
\label{si:sec:relation-to-ewald-mp}
\rebuttal{In this section, we briefly discuss the Ewald-inspired MP formalism proposed in Ref.~\citenum{kosmala2023ewald} and how it relates to EFA. For the $c$-th feature channel, an Ewald MP update is given as}
\begin{equation}
    \rebuttal{\boldsymbol{x}_{mc}^\text{Ewald-MP} \sim \sum_{\vec{k} \in \mathcal{K}} e^{-i \vec{k} \cdot \vec{r}_m} \zeta\Big(\big\lVert\vec{k}\big\rVert_2\Big)_c \, \sum_{n = 1}^N e^{i \vec{k} \cdot \vec{r}_n} \boldsymbol{x}_{nc}\,,} 
    \label{si:eq:ewald-mp}
\end{equation}
\rebuttal{where $\mathcal{K}$ denotes the set of k-vectors $\vec{k} \in \mathbb{R}^3$, constructed from the reciprocal space vectors (derived from the lattice vectors) and $\zeta:\mathbb{R} \mapsto \mathbb{R}^H$ is a trainable filter function in $\vec{k}$-space, e.g., implemented via an MLP. Different k-vectors can be associated with different frequencies in Fourier space such that the sum over $\mathcal{K}$ denotes a sum over different frequencies (implied by $\vec{k}$), which are shared across feature channels.}

\rebuttal{In contrast to EFA, Ewald MP only considers invariant features $\boldsymbol{x}$ and does not include directional information. For this reason, we also restrict ourselves to the EFA formulation with invariant features and additionally omit the feature map $\boldsymbol{\psi}$ in the interest of clarity. This simplified version of EFA can be written as}
\begin{equation}
    \rebuttal{\boldsymbol{x}_{mc}^\text{EFA} \sim \sum_{\hat{u} \in \mathcal{U}} \sum_{d=1}^H \boldsymbol{q}_{md} e^{-i \omega_d \hat{u} \cdot \vec{r}_m} \sum_{n=1}^N e^{i \omega_d \hat{u} \cdot \vec{r}_n} \boldsymbol{k}_{nd}\boldsymbol{v}_{nc}\,,}
    \label{si:eq:simplified-ewald-efa}
\end{equation}
\rebuttal{where $c$ and $d$ denote the feature channels of queries $\boldsymbol{q}$, keys $\boldsymbol{k}$ and values $\boldsymbol{v}$. Fundamental differences between Ewald MP and EFA are the attention-like structure in EFA as well as the summation over frequencies to obtain attention scores, i.e., each feature channel is assigned a specific frequency in EFA. Moreover, Ewald MP includes a trainable filter function not present in EFA, and the summation in Ewald MP runs over k-vectors constructed from the lattice vectors, whereas the sum over $\hat{u}$ in EFA corresponds to a numerical integral over the unit sphere. For periodic systems, it is in principle possible to choose special $\hat{u}$ (instead of an integration grid, see also \nameref{si:sec:other-symmetrisation-operations}) and frequencies $\omega$ for EFA such that $\omega_k \hat{u} \cdot \vec{r} = \vec{k} \cdot \vec{r}$ in the sums $\sum_{\mathcal{K}}$ (\autoref{si:eq:ewald-mp}) and $\sum_{\mathcal{U}} \sum_{d=1}^H$ (\autoref{si:eq:simplified-ewald-efa}). In this special case, simplified EFA and Ewald MP show similar functional forms. However, for non-periodic systems, establishing a similar connection is not possible, as there is no lattice and no associated k-vectors.}

\phantomsection
\section{Molecular Benchmarks Dominated by Local Interactions}\label{si:sec:established-local-benchmarks}
\begin{figure}
    \centering
    \includegraphics[width=0.99\linewidth]{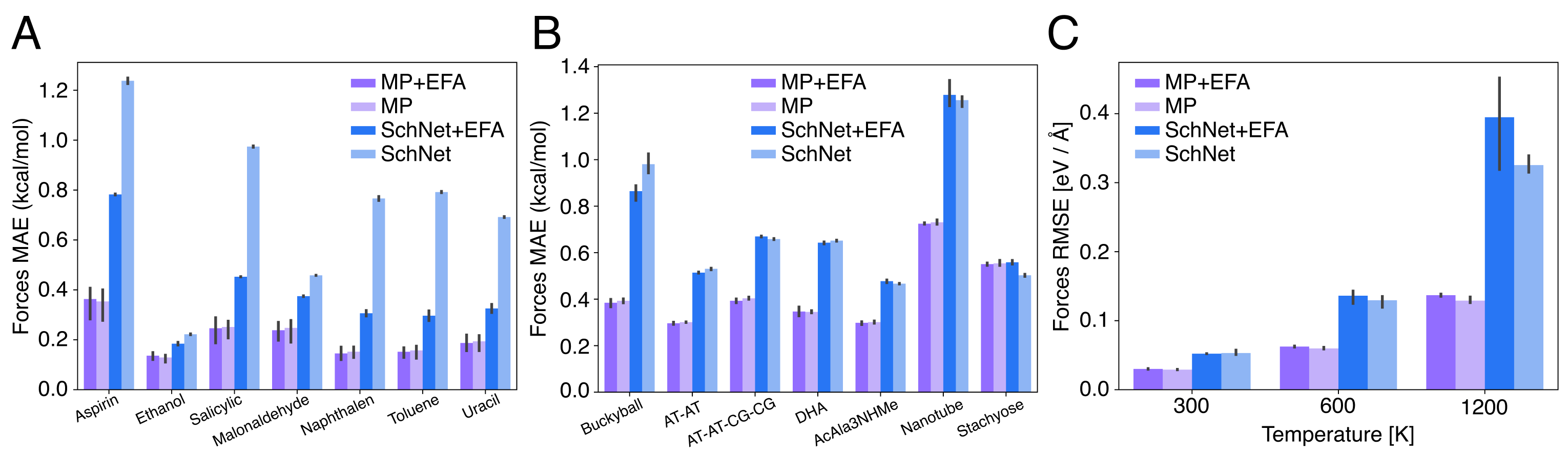}
    \caption{\rebuttal{\textbf{Established, Local Benchmarks.} Results on the established benchmarks for probing local interactions, \textbf{(A)} MD17,\cite{chmiela2017machine} \textbf{(B)} MD22,\cite{chmiela2023accurate} and \textbf{(C)} 3BPA.\cite{kovacs2021linear} As the performance on these benchmarks is dominated by the accuracy of the learned local interactions, there are only small differences between the standard local models and the EFA-augmented counterparts. Results are averaged over three runs with different random seed and error bars denote the $2\sigma$ confidence interval.}}
    \label{si:fig:established-local-benchmarks}
\end{figure}

\rebuttal{Many established MLFF benchmarks mainly probe local interactions, which typically are considerably larger in magnitude than long-range contributions. For this reason, we would not expect any benefit from applying long-range ML methods, and specialised benchmarks (such as the ones investigated in the main body of the text) are required to showcase their utility. However, it is still instructive to benchmark EFA on datasets dominated by short-range interactions: While we would expect no accuracy improvements in this setting, it is important to confirm that augmenting models with EFA does not lead to regressions in their ability to learn short-range interactions. For completeness, we therefore test a standard SchNet\cite{schutt2018schnet} model and the MPNN architecture used throughout this work as well as EFA-augmented variants on three established, local benchmarks: MD17,\cite{chmiela2017machine} MD22,\cite{chmiela2023accurate} and the 3BPA dataset.\cite{kovacs2021linear} We find that standard and EFA-augmented versions yield comparable results on all benchmarks (\autoref{si:fig:established-local-benchmarks}). An exception to this trend is the performance of SchNet on the MD17 benchmark, where including EFA (unexpectedly) leads to significantly lower errors (we did not investigate potential reasons for this observation). For MD17, models are trained on 1k data points following prior work~\cite{schutt2018schnet, batzner20223, frank2022so3krates}, for MD22 we follow the data splits from Ref.~\citenum{chmiela2023accurate} and for 3BPA we use 500 data points for training. For MD17 and MD22, all remaining data is used for validation, and for 3BPA we use the different temperature data splits for testing from the original publication.\cite{kovacs2021linear}}

\begin{figure}
    \centering
    \includegraphics{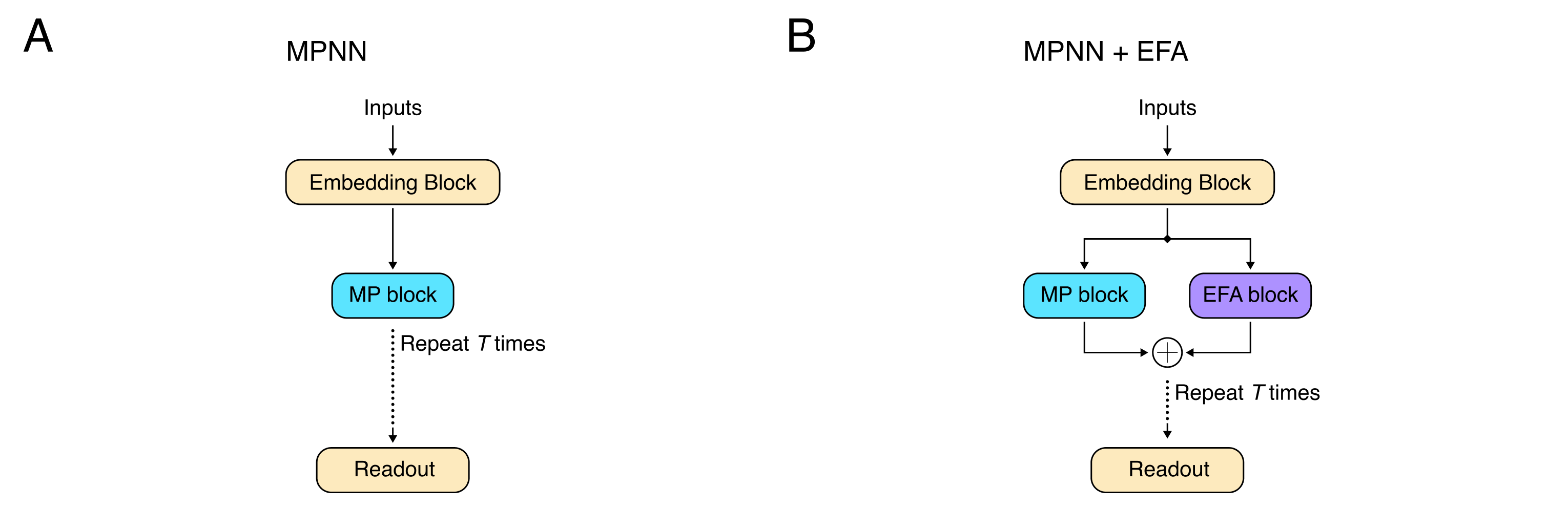}
    \caption{\textbf{Augmenting local message passing (MP) models with Euclidean fast attention (EFA).} \textbf{(A)} Computational flow of a standard MP model vs.~\textbf{(B)} the EFA augmented MP model.}
    \label{si:fig:mp-vs-mp-plus-efa}
\end{figure}

\section{Cusp in Cumulene} \label{si:sec:cusp-in-cumulene}
\rebuttal{As described in the main text, all tested models are unable to reproduce the sharp cusp in the energy profile at a dihedral angle of 180$^\circ$. However, the cusp is an artefact due to the inability of the reference \textit{ab initio} method to describe the non-adiabatic couplings at the conical intersection between two potential energy surfaces. A cusp like this leads to discontinuous forces, which would result in unstable MD simulations. For this reason, most MLFFs (including all models tested here) make discontinuous forces impossible by design: ``Smoothing'' of the problematic region in the energy profile is actually a desirable feature, as it mimics the appearance of a (physically correct) diabatic surface.}

\section{Comparison to Other Models}
The main text focuses mainly on experiments where a basic MPNN architecture is compared to an EFA-augmented variant. For completeness, in this section we repeat some of these experiments with other model architectures from the literature to demonstrate that similar trends can be observed in these cases.

\begin{figure*}
    \centering
    \includegraphics[width=\linewidth]{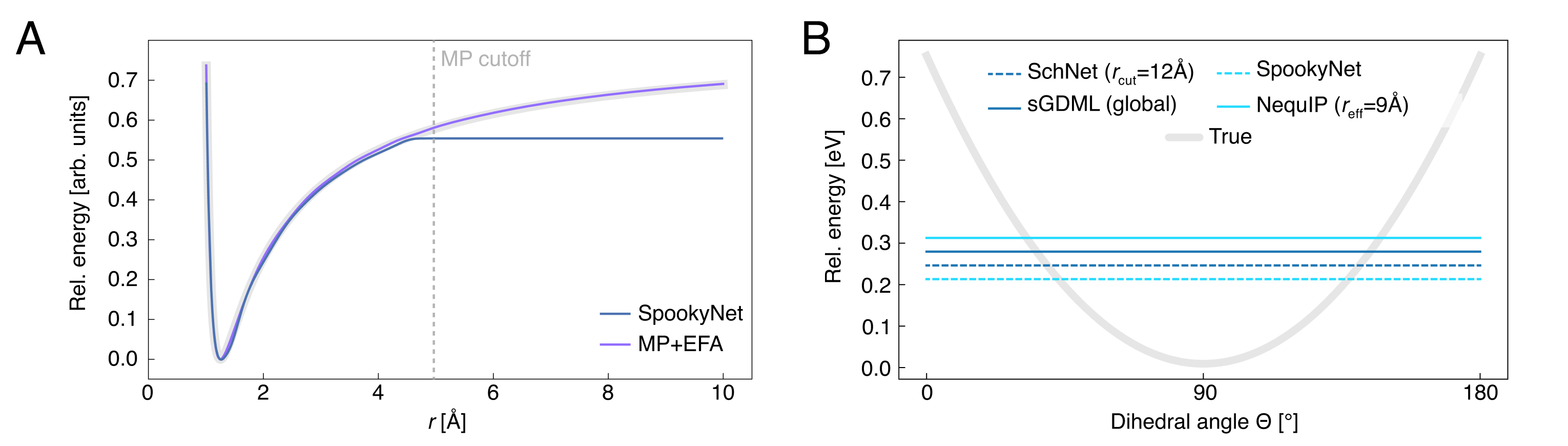}
    \begin{subfigure}{\linewidth}
        \phantomcaption{}
        \label{si:fig:comparison-other-models:pairwise}
    \end{subfigure}
    \begin{subfigure}{\linewidth}
        \phantomcaption{}
        \label{si:fig:comparison-other-models:cumulene}
    \end{subfigure}
    \caption{\textbf{Comparison to other models.} \textbf{(A)} Pairwise potential for $N=2$ atoms system using SpookyNet with linear scaling attention and MP+EFA. \textbf{(B)} Learned energy profile on the cumulene structure using different approaches from the literature.}
    \label{si:fig:comparison-other-models}
\end{figure*}

\paragraph{Pairwise Potential}
For the pairwise potential we train on the two atom system investigated in the main text using a maximal separation of $10~\an$. We compare an MP+EFA model (with the same settings as in the main text) and a SpookyNet model with the default settings from the original paper. The SpookyNet model includes non-local corrections via linear scaling self-attention on the atomic representations. However, this neglects the relative positioning $\vec{r}_m - \vec{r}_n$ of the atoms \wrt each other and consequently cannot describe the potential beyond the local message passing cutoff (\autoref{si:fig:comparison-other-models:pairwise}).
\paragraph{Cumulene} For comparison we choose a global ($r_{\text{cut}} = 12\,\an$) SchNet model, a global (per construction) sGDML\cite{chmiela2019sgdml} kernel, a SpookyNet model, and a NequIP\cite{batzner20223} model with an effective cutoff of 9$\,\an$ (\autoref{si:fig:comparison-other-models:cumulene}). Both SchNet and sGDML rely on invariant descriptors that are based on pairwise distances only. Due to the large separation of the hydrogen rotors, changes in the dihedral angle correspond to extremely small variations in the pairwise distances which are difficult to resolve for these models. The linear self-attention mechanism in SpookyNet fails since it is not able to discriminate interaction patterns which depend on Euclidean information. The equivariant NequIP model is capable of solving the problem by using a sufficient number of MP layers,\cite{frank2022so3krates} but as soon as the effective cutoff becomes too small, it cannot solve the regression task. Simply increasing the number of MP layers to fix this issue is practically infeasible due to the increasing computational cost and information blur (over-squashing).

\section{\texorpdfstring{S$_\text{N}$2 Reactions - SchNet}{SN2 - SchNet}}
For the SchNet model we use the same feature dimensions and follow the implementation from the original publication,\cite{schutt2018schnet} which results in an equal parameter number of 255k for SchNet and SchNet+EFA. The results are shown in \autoref{tab:sn2-schnet}. It highlights the ability of EFA to improve the performance across MPNN backbones.
\begin{table}
    \centering
    \begin{tabular}{lcccc}
        \toprule
        Metric & MP$_{5\an}$ & MP$_{10\an}$ & MP$_{5\an}$+EFA & MP$_{5\an}$+Dispersion \\
        \midrule
        \textit{Energy} & $72.1$ & $38.6$ & $\bm{2.1}$ & $73.1$ \\
        \midrule
        \textit{Forces} & $13.5$ & $3.2$ & $\bm{1.6}$ & $13.6$ \\
        \bottomrule
    \end{tabular}
    \caption{\textbf{Performance on the S$_\text{N}$2 dataset.} Mean absolute errors (MAEs) for energy and forces in meV and meV/$\an$. MAEs are reported with local cutoffs of 5~\an\ (with and without EFA augmentation) and 10~\an\, \rebuttal{and also for an MP model with an analytic dispersion correction (using the functional form of the D2 correction\cite{grimme2006semiempirical} with trainable parameters).} Best model shown in bold.}
    \label{tab:sn2}
\end{table}
\begin{table}
    \centering
    \begin{tabular}{lccccccccc}
        \toprule
        Data set & $r_\text{max}$ & $N_\text{tot}$ & $N_\text{train}$ & $N_\text{valid}$ & $B$ & $N_\text{epochs}$ & $\lambda_E$ & $\lambda_F$ \\
        \midrule
        $k$-chains$_{k = 2}$  & $15\,\an$ & 2 & 2 & 0 & 2 & 1000 & - & -\\
        \midrule
        $k$-chains$_{k = 4}$  & $25\,\an$ & 2 & 2 & 0 & 2 & 1000 & - & -\\
        \midrule
        $k$-chains$_{k = 6}$  & $35\,\an$ & 2 & 2 & 0 & 2 & 1000 & - & -\\
        \midrule
        $k$-chains$_{k = 8}$  & $45\,\an$ & 2 & 2 & 0 & 2 & 1000 & - & -\\
        \midrule
        Pair & $30\,\an$ & 10k & 3500 & 500 & 10 & 1000 & 0.01 & 0.99\\
        \midrule
        Cluster ($d = 10\,\an$) & $30\,\an$ & 1500 & 1000 & 200 & 4 & 100 & $1 / \mathrm{std}_E$ & $10 / \mathrm{std}_{\vec{F}}$\\
        \midrule
        Cluster ($d = 15\,\an$) & $30\,\an$ & 1500 & 1000 & 200 & 4 & 100 & $1 / \mathrm{std}_E$ & $10 / \mathrm{std}_{\vec{F}}$\\
        \midrule
        Cluster ($d = 20\,\an$) & $30\,\an$ & 1500 & 1000 & 200 & 4 & 100 & $1 / \mathrm{std}_E$ & $10 / \mathrm{std}_{\vec{F}}$\\
        \midrule
        Cluster ($d = 25\,\an$) & $30\,\an$ & 1500 & 1000 & 200 & 4 & 100 & $1 / \mathrm{std}_E$ & $10 / \mathrm{std}_{\vec{F}}$\\
        \midrule
        Cluster ($d = 30\,\an$) & $30\,\an$ & 1500 & 1000 & 200 & 4 & 100 & $1 / \mathrm{std}_E$ & $10 / \mathrm{std}_{\vec{F}}$\\
        \midrule
        BIGDML - Graphene & - & 20538 & 200 & 50 & 4 & 5000 & 0.01 & 0.99\\
        \midrule
        BIGDML - Na & - & 9066 & 200 & 50 & 4 & 5000 & 0.01 & 0.99\\
        \midrule
        BIGDML - Pd$_32$H & - & 6772 & 200 & 50 & 4 & 5000 & 0.01 & 0.99\\
        \midrule
        BIGDML - PdMgO & - & 19098 & 200 & 50 & 4 & 5000 & 0.01 & 0.99\\
        \midrule
        BIGDML - Pd\,(1000K) & - & 12024 & 200 & 50 & 4 & 5000 & 0.01 & 0.99\\
        \midrule
        BIGDML - Pd\,(500K) & - & 7757 & 200 & 50 & 4 & 5000 & 0.01 & 0.99\\
        \midrule
        4GHDNNP - Au$_2$MgO & - & 5000 & 4500 & 500 & 8 & 2000 & 0.01 & 0.99\\
        \midrule
        4GHDNNP - Carbon Chain & $15\,\an$ & 10000 & 9000 & 500 & 8 & 500 & 0.01 & 0.99\\
        \midrule
        4GHDNNP - NaCl & $15\,\an$ & 5000 & 4500 & 500 & 8 & 500 & 0.01 & 0.99\\
        \midrule
        Charge-Dipole & $10\,\an$ & 10k & 2500 & 500 & 10 & 3000 & 0.01 & 0.99 \\
        \midrule
        S$_\text{N}$2 & $20 \,\an$ & 452k & 405k & 5000 & 32 & 500 & 0.01 & 0.99 \\
        \midrule
        Cumulene & $15\,\an$ & 4973 & 1500 & 500 & 5 & 2000 & 0.01 & 0.99 \\
        \midrule
        Dimer & $15\,\an$ & 4612 & 4500 & 250 & 16 & 6000 & 0.50 & 0.50 \\
        \bottomrule
    \end{tabular}
    \caption{\textbf{Summary of datasets and training hyperparameters.} The columns denote the maximal separation in the data set between atoms $r_\text{max}$ rounded to the next highest multiple of 5 (this value is used in \autoref{eq:max-frequency-bound} to determine the maximal frequency), the total number of data points $N_\text{tot}$ in the data set, the total number of points used for training $N_\text{train}$ including the number of points $N_\text{valid}$ used for validation to select the best model during training, the batch size $B$, the number of epochs $N_\text{epochs}$, and the trade-off parameters for energy $\lambda_E$ and forces $\lambda_F$ in the loss function. \rebuttal{Note that for periodic systems, no $r_\text{max}$ is specified, since no integration over the unit sphere is performed. For the Cluster data (\autoref{fig:results-panel-NaCl_toy}), we scale the energy and force loss weights with the inverse of the standard deviation (std) over the training data. This ensures an effective equal weighting between energy and forces for the different training runs. We further chose the same $r_\text{max}$ for all clusters, to maintain constant maximal frequency in EFA (see \autoref{sec:methods}).} As $k$-chains is a classification task we train it via softmax cross-entropy loss.}
    \label{si:tab:datasets-and-training}
\end{table}
\begin{table}[t]
    \centering
    \begin{tabular}{ccccccccccccccc}
    \toprule
    $\numgridpoints$ & 50 & 86 & 110 & 146 & 194 & 230 & 266 & 302 & 350 & 434 & 590 & 770 & 974 & 6000
    \\
    \midrule
    $b_\text{max}$ & $\pi$ & $2\pi$ & $2.5\pi$ & $3\pi$ & $4\pi$ & $4.5\pi$ & $5\pi$ & $5.5\pi$ & $6.5\pi$ & $7.5\pi$ & $9\pi$ & $11\pi$ & $12.5\pi$ & $35\pi$
    \\
    \bottomrule
    \end{tabular}
    \caption{\textbf{Grid size for Lebedev quadrature vs.~maximal value for $b_\text{max}$.} See \autoref{eq:max-frequency-bound} from the \nameref{sec:methods} section in the main text for details. Values have been determined via comparison to the analytic solution for spherical harmonics with $L_Y = 0$. Note that for $L_Y > 0$, the numerical precision of the quadrature decreases slightly (also shown in \autoref{si:fig:efa-vs-bessel}).}
    \label{si:tab:bmax-to-lebedev-num}
\end{table}

\begin{figure*}
    \centering
    \includegraphics[width=\linewidth]{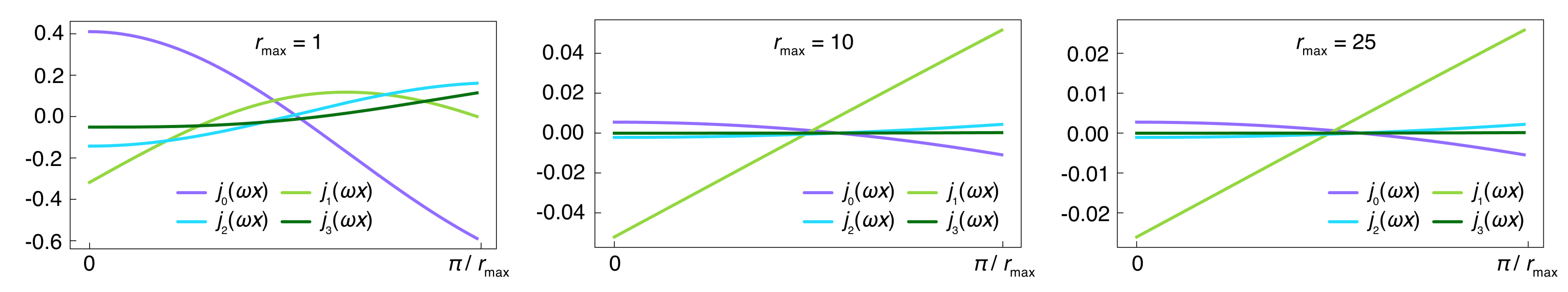}
    \caption{\textbf{Radial Bessel function for varying maximal distance.} Shape of the Bessel function of different degree $\ell$ and varying value of $r_\text{max}$, which changes the maximal frequency value $\omega_\text{max}$ as described in the \nameref{sec:methods} section (\autoref{eq:max-frequency-bound}).}
    \label{si:fig:bessel-rescaling}
\end{figure*}

\begin{figure}
    \centering
    \includegraphics[width=\linewidth]{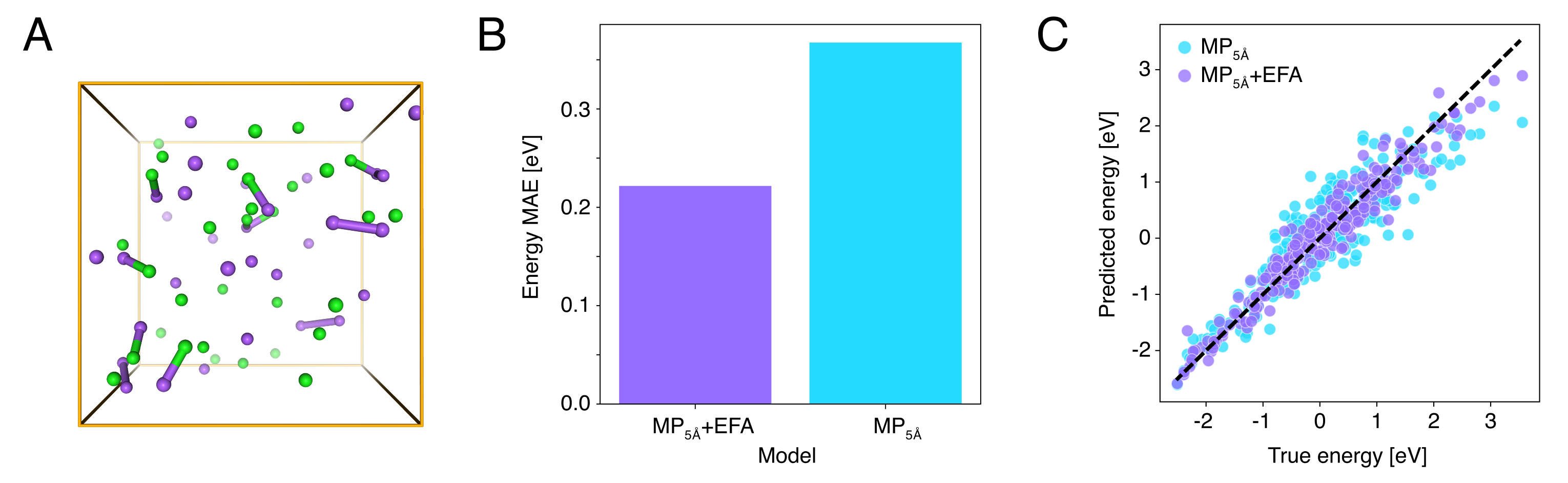}
    \caption{\rebuttal{\textbf{NaCl bulk system.} \textbf{(A)} Visualisation of the periodic NaCl system proposed in Ref.~\citenum{dumittan2023lode}. \textbf{(B)} Mean absolute error (MAE) for energy for a standard message passing (MP) model and an EFA-augmented version. \textbf{(C)} Correlation between the ground truth energies and the energies predicted by MP and MP+EFA (perfect correlation is indicated by the dotted black line). Since the data set contains different cell sizes, we employ normalized lattice vectors in \autoref{eq:pbc_symmetrisation}.}}
    \label{si:fig:NaCl-bulk}
\end{figure}
\begin{figure}
    \centering
    \includegraphics[width=\linewidth]{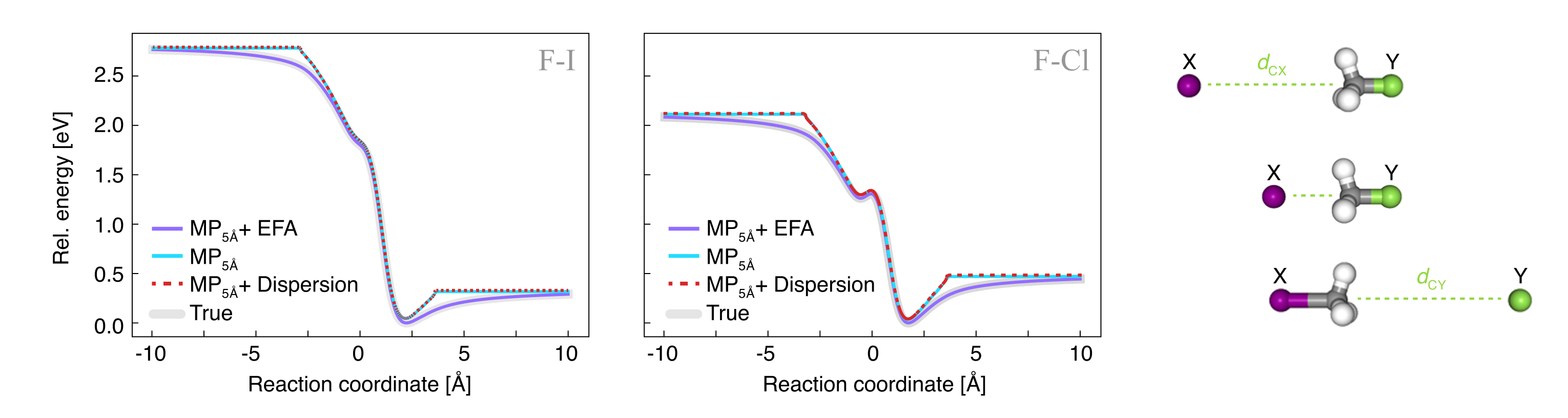}
    \phantomcaption{}
    \label{si:fig:sn2-with-dispersion}
    \caption{\rebuttal{\textbf{S$_\text{N}$2 reactions with dispersion correction.} Minimum energy path for a message passing (MP) model, an MP model augmented with Euclidean fast attention (EFA), and an MP model with D2 dispersion correction,\cite{grimme2006semiempirical} where the C6 coefficients are fully trainable. Using dispersion corrections does not allow to correctly model the energy profile, because dispersion interactions decay as $r^{-6}$, whereas the relevant long-range interactions in the S$_\text{N}$2 dataset decay as $r^{-2}$. This underlines that domain knowledge is needed to fix the long-range behaviour with analytical corrections, whereas EFA can be applied without any physical prior knowledge.}}
\end{figure}
\begin{figure*}
    \centering
    \includegraphics[width=\linewidth]{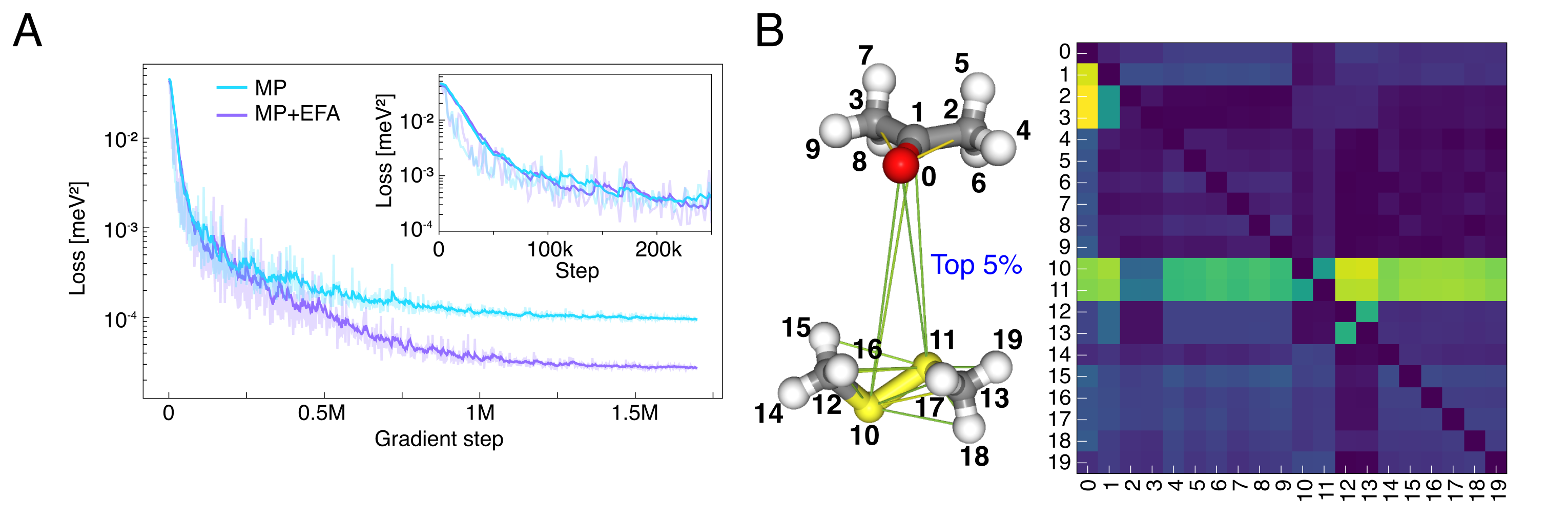}
    \begin{subfigure}{\linewidth}
        \phantomcaption{}
        \label{si:fig:training-dynamics-attention-analysis:loss}
    \end{subfigure}
    \begin{subfigure}{\linewidth}
        \phantomcaption{}
        \label{si:fig:training-dynamics-attention-analysis:attention}
    \end{subfigure}
    \caption{\textbf{Training dynamics and attention analysis. }\textbf{(A)} Loss as a function of gradient step for MP and MP+EFA on the dimer data set. Inset shows the loss over the first 250k steps. \textbf{(B)} Visualisation of the learned attention map for a randomly selected dimer. In the 3D view, the pairwise attention values which belong to the largest 5\% in magnitude are shown.}
    \label{si:fig:training-dynamics-attention-analysis}
\end{figure*}
\begin{figure*}
    \centering
    \includegraphics[width=\linewidth]{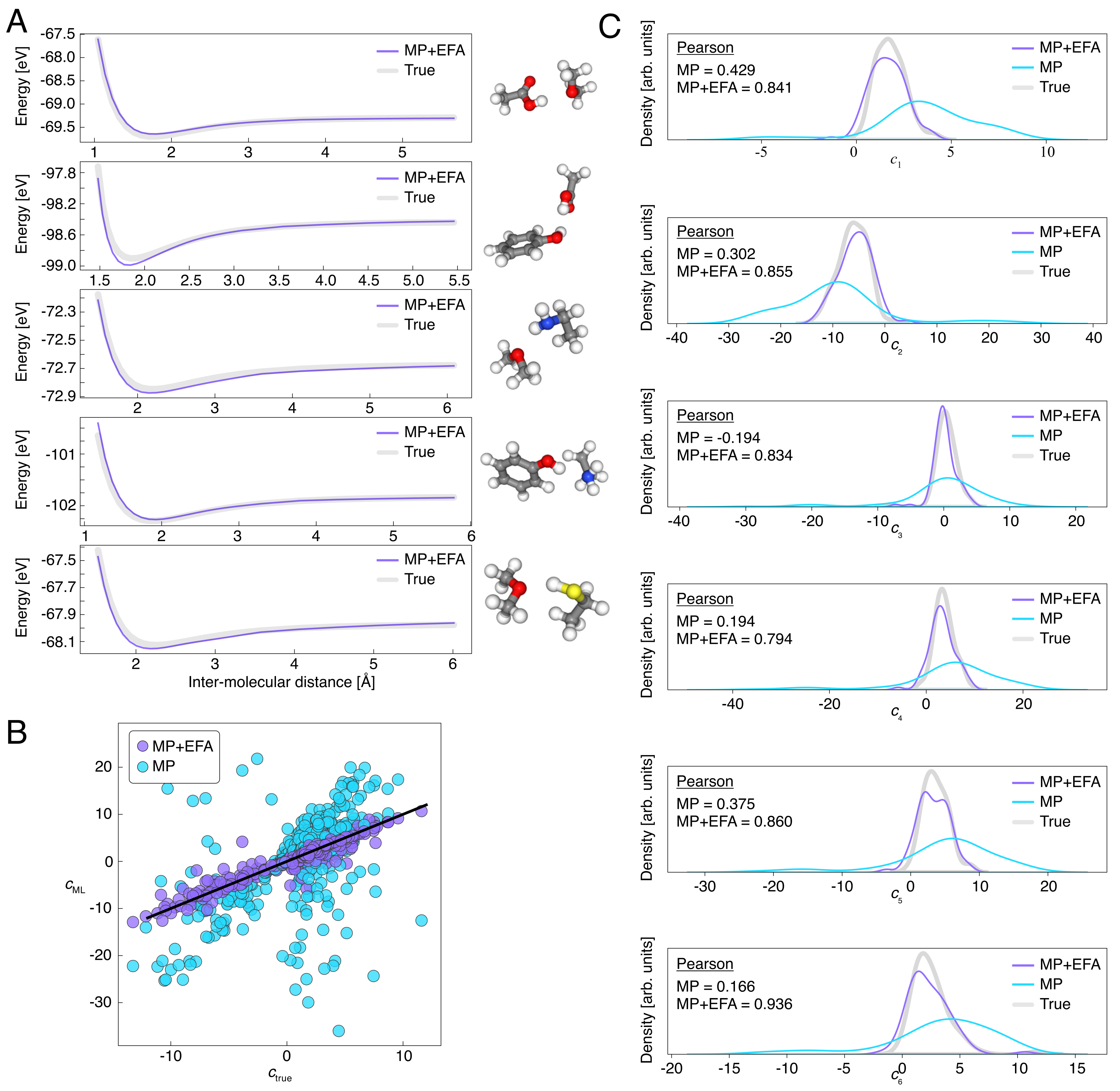}
    \begin{subfigure}{\linewidth}
        \phantomcaption{}
        \label{si:fig:dimer:potential}
    \end{subfigure}
    \begin{subfigure}{\linewidth}
        \phantomcaption{}
        \label{si:fig:dimer:correlation}
    \end{subfigure}
    \begin{subfigure}{\linewidth}
        \phantomcaption{}
        \label{si:fig:dimer:distribution}
    \end{subfigure}
    \caption{\textbf{Analysis of dimer experiments.} \textbf{(A)} Energy profile for five completely unknown dimers that have not been part of the training data. \textbf{(B)} Scatter plot of the coefficients fitted to the long range tail of the true energy profiles vs.~the coefficients fitted to the prediction of the message passing (MP) model and the MP + Euclidean fast attention (EFA) model. \textbf{(C)} Individual distribution for each fitted coefficient $c_1, \dots, c_6$. The Pearson correlation coefficients between the coefficients fitted to the true energy profile and the MP and MP+EFA models are reported in each panel.}
    \label{si:fig:dimer}
\end{figure*}
\begin{table}[t]
    \centering
    \begin{tabular}{lccc}
        \toprule
        & SchNet$_{5\an}$ & SchNet$_{10\an}$ & SchNet$_{5\an}$+EFA \\
        \midrule
        MAE$_\text{Energy}$ & $76.3\,(\pm 0.0)$ & $38.7\,(\pm 0.7)$ & $\bm{2.0\,(\pm 0.2)}$ \\
        \midrule
        MAE$_\text{Forces}$ & $18.1\,(\pm 0.1)$ & $3.7\,(\pm 0.1)$ & $\bm{1.8\,(\pm 0.1)}$ \\
        \bottomrule
    \end{tabular}
    \caption{\textbf{Performance on the S$_\text{N}$2 dataset for SchNet.} Mean absolute errors (MAEs) for energy and forces in meV and meV/$\an$. MAEs are reported for SchNet with local cutoffs of 5~\an\ (with and without EFA augmentation) and 10~\an\ (without EFA augmentation). The term in brackets denotes the standard deviation over three different training runs (with different random seeds). Best model shown in bold.}
    \label{tab:sn2-schnet}
\end{table}

\end{document}